\documentclass[11pt]{article}

\usepackage{latexsym,mathrsfs}
\usepackage{amssymb,amsmath} 
\usepackage{amsthm,enumerate,verbatim}
\usepackage{amsfonts}
\usepackage{graphicx}
\usepackage{algorithm}
\usepackage{algorithmic}
\usepackage{url}
\usepackage{color}
\usepackage{caption}
\usepackage{subcaption}
 \usepackage{rotating} 
\usepackage{multirow}
\usepackage{amsopn}
\usepackage{times}

\usepackage{rotating}
\usepackage{bbm}
\usepackage{latexsym}

\usepackage{lipsum}
\usepackage{epstopdf}

\newtheorem{lemma}{Lemma}

\newtheorem{theorem}{Theorem}

\newtheorem*{definition*}{Definition}
\newtheorem{example}{Example}

\newtheorem{note}{Note}

\newcommand{\mcalX}{\mathcal{X}}

\DeclareMathOperator{\argmin}{argmin} 
\DeclareMathOperator{\cone}{cone} 
\DeclareMathOperator{\rank}{rank}

\DeclareMathOperator{\logdet}{logdet}

\definecolor{brightpink}{rgb}{1.0, 0.0, 0.5}

\newcommand{\revises}[1]{{{\color{black} #1}}}

\author{Valentin Leplat\thanks{Center for Artificial Intelligence Technology,
Skoltech, Moscow, Russia
  ({V.Leplat@skoltech.ru}).}  
\and Le Thi Khanh Hien\thanks{Department of Mathematics and Operational Research, University of Mons, Mons, Belgium. NG and LTKH acknowledge the support by the European Union (ERC consolidator, eLinoR, no 101085607). 
  ({khanhhiennt@gmail.com,nicolas.gillis@umons.ac.be).}}
  \and Akwum Onwunta\thanks{Department of Industrial and Systems Engineering,
Lehigh University, Bethlehem, USA
  ({ako221@lehigh.edu}).}
\and Nicolas Gillis\footnotemark[3]}

\title{Deep Nonnegative Matrix Factorization \\ with  Beta Divergences} 
\date{}

\newcommand{\revise}[1]{{{\color{black} #1}}}

\begin{document}

\maketitle

\begin{abstract}
Deep Nonnegative Matrix Factorization (deep NMF) has recently emerged as a valuable technique for extracting multiple layers of features across different scales. However, all existing deep NMF models and algorithms have primarily centered their evaluation on the least squares error, which may not be the most appropriate metric for assessing the quality of approximations on diverse datasets. For instance, when dealing with data types such as audio signals and documents, it is widely acknowledged that $\beta$-divergences offer a more suitable alternative. In this paper, we develop new models and algorithms for deep NMF using some $\beta$-divergences, with a focus on the Kullback-Leibler divergence. Subsequently, we apply these techniques to the extraction of facial features, the identification of topics within document collections, and the identification of materials within hyperspectral images.  
\end{abstract}


\section{Introduction}

\begin{figure}[ht!]
\begin{center}
\begin{tabular}{ccc}
     \includegraphics[height=6cm]{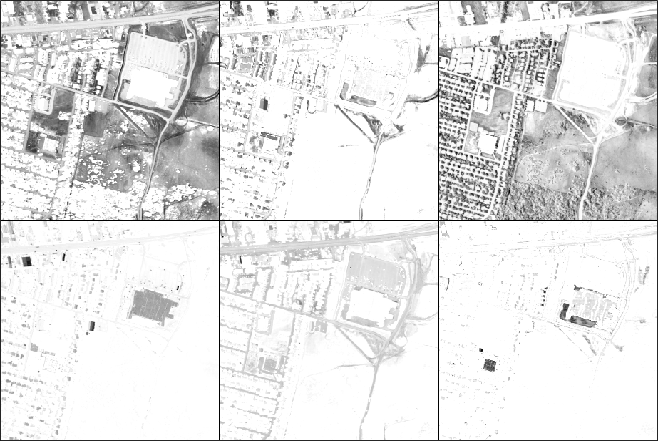} & \quad & 
      \includegraphics[height=6cm]{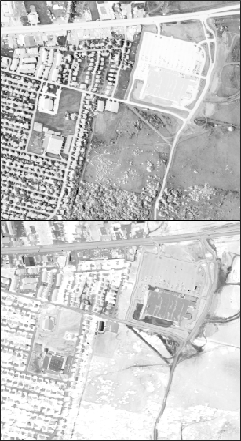} \\ 
      (a) & & (b) 
\end{tabular} 
\caption{Deep NMF applied on the Urban hyperspectral image, which is an aerial image of a Walmart in Copperas Cove, Texas. We can for example easily identify the roof top and the parking lot of the store; see the fourth and fifth image in (a), respectively.  
Using Deep NMF with two layers, we obtain the following: (a)~Layer~1 with $r_1 = 6$ contains the abundance maps $H_1$ corresponding to the spectral signatures in $W_1$, 
and (b)~Layer~2 with $r_2 = 2$ contains the abundance maps $H_2 H_1$ corresponding to the spectral signatures in $W_2$.  
As the factorization unfolds, deep NMF generates denser abundance maps which are combinations of abundance maps from previous layers. Here, the first level extracts 6 materials (including grass, roof tops and dirt, trees, other roof tops, road and dirt), which are merged into vegetation vs.\ non-vegetation at the second layer.}  
\label{fig:deepnmfurban}
\end{center}
\end{figure}

Deep NMF seeks to approximate an input data matrix $X \in \mathbb{R}^{m \times n}_+$, as follows: 
\begin{equation}\label{eq:modelLC}
X \approx W_1 H_1, \quad 
W_1 \approx W_2 H_2, \quad 
\dots \quad 
W_{L-1} \approx W_L H_L, 
\end{equation}
where $W_\ell \in \mathbb{R}^{m \times r_\ell}_+$ 
and 
$H_\ell \in \mathbb{R}^{r_{\ell} \times r_{\ell-1}}_+$ \revise{are the factors of the $\ell$th layer matrix factorization, $W_{\ell-1} \approx W_\ell H_\ell$}, for $\ell = 
1,2,\dots,L$, and with $r_0 = n$.  This approach yields a total of $L$ layers of decompositions for $X$:
\begin{equation}\label{eq:modelDC}
    X \approx W_1 H_1, \quad 
X \approx W_2 H_2 H_1, \quad 
\dots, \quad 
X \approx W_L H_L H_{L-1} \dots H_1.
\end{equation}
For example, let $X \in \mathbb{R}^{m \times n}_+$ represent a hyperspectral image, with $m$ spectral bands and $n$ pixels, where $X(:,j)$ is the spectral signature of the $j$th pixel and $X(i,:)$ is the vectorized image corresponding to the $i$th wavelength. 
Then the first layer of the factorization, $X \approx W_1 H_1$, is such that 
$W_1 \in \mathbb{R}^{m \times r_1}_+$ contains the spectral signatures of $r_1$ materials, while $H_1  \in \mathbb{R}^{r_1 \times n}_+$ contains the so-called abundance maps that indicate which material is present in which pixel and in which proportion.  Figure~\ref{fig:deepnmfurban}~(a) provides an example where the extracted materials include  grass, trees, roof tops, and roads. 
This is the same interpretation as for NMF~\cite{ma2013signal}.  
At the second layer, $W_1 \approx W_2 H_2$ so that  
$W_2 \in \mathbb{R}^{m \times r_2}_+$ contains the spectral signatures of higher-level materials; for example vegetation vs.\ non-vegetation in Figure~\ref{fig:deepnmfurban}~(b). In other words, $W_2$ will merge similar materials in a single material (e.g., grass and trees in the vegetation). The factor $H_2 \in \mathbb{R}^{r_2 \times r_1}$ indicates how the low-level materials are combined into high-level materials. 


 We will make the assumption that the ranks decrease as the factorization proceeds, specifically, that $r_{\ell+1} < r_\ell$ for all $\ell$. This rank reduction is the most natural and common scenario. It is important to note that employing $r_{\ell+1} > r_\ell$ leads to overparametrization, which can have its merits in certain contexts, such as cases involving implicit regularization, as discussed in~\cite{arora2019implicit}. However, our primary objective in this paper is not to pursue overparametrization. 

There has been a recent surge of research on deep NMF. 
It began with the pioneering work of~\cite{cichocki2006multilayer, cichocki2007multilayer}, which focused on multilayer NMF techniques that sequentially decompose the input matrix $X$: \revises{it first decomposes $X = W_1H_1$ using NMF, then $W_1=W_2H_2$, then $W_2=W_3H_3$ and so on,  following~\eqref{eq:modelLC}. As it is sequential, Multilayer NMF does not consider a global optimization problem trying to balance the errors between layers.}   
Subsequently, \cite{trigeorgis2014deep, trigeorgis2016deep} introduced deep NMF, presenting a model closely related to the formulation in~\eqref{eq:modelDC}.  \revises{Deep NMF first performs multilayer NMF but then reoptimizes the factors at all layers iteratively; see Section~\ref{sec:model}.}. 
Deep NMF has found applications across a diverse range of fields, including recommender systems~\cite{mongia2020deep}, community detection~\cite{ye2018deep}, and topic modeling~\cite{will2023neural}. For more comprehensive insights and surveys on deep matrix factorizations, readers can refer to~\cite{de2021survey} and~\cite{chen2022survey}. These surveys offer up-to-date overviews of the field and its recent advancements. To the best of our knowledge, it is noteworthy that all existing deep NMF models employ the Frobenius norm, which corresponds to the least squares error, as the standard metric to assess the reconstruction error at each layer. Furthermore, a significant portion of the literature has tended to overlook the modeling aspects inherent to deep NMF. Consequently, many studies have adopted inconsistent models throughout the different layers of the factorization process, as highlighted in~\cite{de2023consistent}, \revise{that is, that the objective function optimized at different layers is not the same preventing a meaningful model (because parameters are optimized following different models) and convergence guarantees; see Section~\ref{sec:model} for more details.} 



\paragraph{Contribution and outline of the paper} 
In this paper, we first focus on the modeling aspect of deep NMF in Section~\ref{sec:model}.  
We explain how to use meaningful regularizations and why a layer-centric loss function is more appropriate than a data-centric one when it comes to identifiability (\revise{that is, the uniqueness of the solution up to scaling and permutation
ambiguities}). This was observed experimentally in~\cite{de2023consistent} but not justified from a theoretical viewpoint. 
Then, in Section~\ref{sec:algorithm}, we propose new regularized models for deep NMF based on $\beta$-divergences, consistent across the layers, and design algorithms for solving the proposed deep NMF models, with a focus on the Kullback-Leibler divergence ($\beta = 1$). As a by-product, we will provide multiplicative updates (MU) for a problem of the type 
    \[
    \min_{W \geq 0} \; D_\beta(X, WH) + \lambda D_\beta(W, \bar{W}), 
    \] 
    where $X$, $H$ and $\tilde{W}$ are fixed, $\lambda$ is a positive penalty parameter,  and $D_\beta$ is a $\beta$-divergence \revise{with $\beta \in \{0, 0.5, 1, 1.5, 2 \}$}. 
Finally, in Section~\ref{sec:appli}, we use the newly proposed models and algorithms 
for facial feature extraction, 
topic modeling, and the identification of materials within hyperspectral images. 


\section{What Deep NMF model to use?} \label{sec:model}

\cite{de2023consistent} introduced two distinct loss functions specifically designed for deep NMF: 
\begin{enumerate}
    \item A data-centric loss function (DCLF) defined as 
\begin{equation}
\mathcal{L}(H_1, H_2,\dots, H_L; W_1, W_2,\dots, W_L) 
= 
\sum_{\ell=1}^L \lambda_\ell 
D\Big( X, W_\ell \prod_{i=0}^{\ell-1} H_{\ell-i} \Big), 
    \label{eq:lossDCLF}
\end{equation} 
 where $D(A,B)$ is a measure of distance between $A$ and $B$, and the $\lambda_\ell$'s are positive penalty parameters. 
 This loss function minimizes a weighted sum of the errors in the decompositions of $X$ at each of the $L$ \revise{layers}, as given in~\eqref{eq:modelDC}.  

\item A layer-centric loss function (LCLF) defined as 
\begin{equation}
\mathcal{L}(H_1, H_2,\dots, H_L; W_1, W_2,\dots, W_L) 
= 
\sum_{\ell=1}^L \lambda_\ell D(W_{\ell-1}, W_\ell H_\ell),  
    \label{eq:lossLCLF}
\end{equation} 
where $W_0 = X$. 
This loss function minimizes a weighted sum of the errors at each layer, as given in~\eqref{eq:modelLC}.   
\end{enumerate} 
In the majority of previous works, the loss function minimized at each layer was not consistent across layers, \revise{that is, the optimization of each factor does not come from the same (global) objective function; see Table~\ref{tab:table1} for an example of a 3-layer model used in~\cite{trigeorgis2016deep}.  
\begin{table}[!h]
\centering
\begin{tabular}{|c | c | c |} 
 \hline
 Layer $l$ & Objective function for $H_l$ & Objective function for $W_l$ \\
 \hline\hline
 1 & $\|X-W_2H_2\mathbf{H_1}\|_F^2$ & $\|X-\mathbf{W_1}H_1\|_F^2$\\
 2 & $\|X-W_3H_3\mathbf{H_2}H_1\|_F^2$  & $\|X-\mathbf{W_2}H_2H_1\|_F^2$\\
 3 & $\|X-W_3\mathbf{H_3}H_2H_1\|_F^2$ & $\|X-\mathbf{W_3}H_3H_2H_1\|_F^2$\\
 \hline
 \end{tabular}
    \caption{Objective functions minimized in~\cite{trigeorgis2016deep} for each factor for $L=3$. Table adapted  from~\cite{de2023consistent}.}
    \label{tab:table1}
\end{table}}     
As a result, their algorithms generally did not converge and often exhibited poorer performance compared to the loss functions introduced in~\cite{de2023consistent}.


\paragraph{Layer centric vs.\ data centric}  
It was empirically observed in~\cite{de2023consistent} that  LCLF (as defined in~\eqref{eq:lossLCLF}) outperformed DCLF (as defined in~\eqref{eq:lossDCLF}) and the state of the art. This superior performance was observed across various synthetic and real datasets, and it extended to the recovery of ground truth factors. Interestingly, the reason behind this performance is not a matter of chance but \revises{can be understood via identifiability considerations}, as indicated in the subsequent sections. 
This insight helps provide a better understanding of the observed empirical results and offers a rationale for the preference of LCLF in the context of deep NMF.

\subsection{Identifiability of NMF} 

One of the primary reasons why LCLF is found to be more efficient than DCLF lies in the fact that LCLF possesses better identifiability properties compared to DCLF (see Section~\ref{sec:ident_deepNMF}). To better understand this distinction,  it is beneficial to revisit some of the key NMF identifiability results. 

\paragraph{The sufficiently scattered condition}
A nonnegative matrix $H \in \mathbb{R}^{r \times n}_+$ satisfies the sufficiently scattered condition (SSC)\footnote{There are several definitions of the SSC, see the discussion in~\cite[Chap.~4]{gillis2020nonnegative}, and we choose here the simplest from~\cite{lin2015identifiability}.} 
if 
\[
\mathcal{C} 
= \{ x \in \mathbb{R}^r_+ 
\ | \ \|x\|_1 \geq q \| x \|_2 \} 
\quad 
\subseteq 
\quad  
\cone(H) = \{ x \ | \ x = Hy, y \geq 0\}, 
\] 
for some $q < \sqrt{r-1}$. 
The set $\mathcal{C}$ is the intersection of the nonnegative orthant with a second-order cone. The SSC implies that $H$ is sufficiently sparse, in particular it requires $H$ to have at least $r-1$ zeros per row; see the discussions in~\cite{fu2019nonnegative} and \cite[Chap.~4]{gillis2020nonnegative}, and the references therein. 
Based on the SSC, we have the following identifiability result for NMF. 
\begin{theorem}\cite{huang2013non} 
\label{th:uniqNMF}
    Let $X = W^* H^*$ be a rank-$r$ NMF of $X$, where  ${W^*}^\top$ and $H^*$ satisfy the SSC. 
    Then any other rank-$r$ NMF of $X$, $X = WH$, corresponds to $(W^*,H^*)$, up to permutation and scaling of the columns of $W^*$ and rows of $H^*$. 
\end{theorem}

Imposing the SSC on both factors, $W$ and $H$, can sometimes be overly restrictive. For instance, in hyperspectral imaging, it makes sense for $H$ to have this constraint because its rows often represent sparse abundance maps. However, assuming the SSC to $W$ is typically not appropriate since it is expected to be dense in many cases. To address this limitation and provide more flexibility, researchers have introduced regularized NMF models. Among these, the minimum-volume NMF is one of the most  effective approaches, both from a theoretical perspective and in practical applications. 

\paragraph{Minimum-volume NMF}  Minimizing the volume of the columns of $W$ is a popular and powerful NMF regularization technique. The most prevalent form of this regularization is achieved by utilizing $\logdet (W^\top W)$ under normalization constraints on either $W$ or $H$, see e.g., \cite{fu2019nonnegative} and \cite[Chap.~4]{gillis2020nonnegative}. This leads to  identifiability/uniqueness of NMF, as stated in Theorem~\ref{th:identifminvolNMF}. In practice, we use $\logdet (W^\top W + \delta I)$ (with the addition of a small parameter $\delta$) for numerical stability; see the discussion in~\cite{leplat2019minimum}.  

\begin{theorem}[\cite{fu2015blind, fu2018identifiability, leplat2020blind}]
\label{th:identifminvolNMF}
    Let $X = W^* H^*$ be a rank-$r$ NMF of $X$, where $\rank(W^*) = r$ and $H^*$ satisfies the SSC. 
    Then any optimal solution of the following problem 
\begin{align*}
\min_{W} \det(W^\top W) \quad  \text{ s.t }\quad   X = WH 
\quad &\text{ and }   H^\top e = e  \text{ \cite{fu2015blind}}
\\&\text{ or }
He = e \text{ \cite{fu2018identifiability}}
\\&
\text{ or }  
W^\top e = e \text{ \cite{leplat2020blind}}, 
\end{align*}
\revise{where $e$ denotes the vector of all ones of appropriate dimension, }  corresponds to $(W^*,H^*)$, up to permutation and scaling of the columns of $W^*$ and rows of $H^*$. 
\end{theorem}
Minimizing the volume of the columns of $W$ also has several important implications:
\begin{itemize}
    \item By encouraging the columns of $W$ to be closer to the data points, this regularization enhances the interpretability of the features represented by these columns. 
    
    \item The regularization leads to sparser factors $H$. When the columns of $W$ are close to the data points, it implies that more data points are located near the faces of the cone generated by these columns. Consequently, this results in a sparser representation of the data in the factor  $H$, where many elements are driven towards zero. 
    
    \item In scenarios where the factorization rank has been overestimated, min-vol NMF can perform automatic rank detection by setting some of the rank-one factors to zero~\cite{leplat2019minimum}. 
\end{itemize}

\subsection{Discussion on identifiability of regularized deep NMF} 
\label{sec:ident_deepNMF}

Regularization plays a crucial role in enhancing the interpretability and identifiability of deep NMF, similar to its importance in standard NMF, as illustrated by Theorem~\ref{th:identifminvolNMF}. When designing a deep NMF model, careful consideration should be given to which factor should be regularized and how it should be done. In the context of LCLF, where $W_\ell$ is factorized at each layer, it is important to note that overly sparse $W_\ell$ matrices can be challenging to approximate with NMF.  For instance, the identity matrix, which is very sparse, has a unique NMF representation of maximum size ($I = I \cdot I$). Therefore, it makes sense to focus on minimizing the volume of $W_\ell$ matrices and/or maximizing the sparsity of $H_\ell$ matrices since this will generate denser $W_\ell$ matrices.

By the aforementioned reasons, adopting the min-vol NMF approach is a reasonable choice to establish a baseline regularization for deep NMF. In the context of LCLF, at each layer, min-vol LCLF aims to find the solution with the minimum volume for the corresponding $W_\ell$. However, applying Theorem~\ref{th:identifminvolNMF} to each layer individually is not possible because it would require the $W_\ell$ matrices to have full rank, which is precluded by construction due to the hierarchical structure where $W_{\ell-1} = W_{\ell} H_{\ell}$ and the assumption $r_\ell < r_{\ell-1}$.  Fortunately, empirical observations suggest that min-vol NMF can recover $W$ even when it is rank-deficient, provided that $H$ is sufficiently sparse, as demonstrated in~\cite{leplat2019minimum}. Additionally, the literature includes sparse NMF models, such as those discussed in~\cite{abdolali2021simplex}, which offer identifiability even in the rank-deficient case. These observations underscore the adaptability and effectiveness of min-vol regularization in various settings within deep NMF, despite rank-deficiency challenges. We therefore have the following intuition:
\emph{when the sparsity of $H_\ell$ is sufficient, min-vol deep NMF employing the LCLF should exhibit identifiability, provided that each layer, $W_{\ell-1} = W_{\ell} H_{\ell}$, is  identifiable. 
}

In the context of DCLF, achieving identifiability  necessitates that the products $\prod_{\ell=1}^p H_\ell$ are sufficiently sparse for each layer, where $p$ ranges from 1 to $L$. However, it is crucial to recognize that the product of sparse nonnegative matrices tends to be denser than the individual factors. Consequently, it becomes significantly less likely for the product of these matrices to exhibit sparsity, which is essential for DCLF to be identifiable.
On the other hand, a necessary condition for the product $H = H_1 H_2$ with $H_1 \geq 0$ and $H_2\geq 0$ to satisfy the SSC is that $H_2$ satisfies the SSC, because $\cone(H) \subseteq \cone(H_2)$. Remarkably, even when both $H_1$ and $H_2$ individually satisfy the SSC, it remains rather unlikely for their product $H_2 H_1$ to satisfy the SSC.
In fact, for $H_2 H_1$ to satisfy the SSC, $H_1$ and $H_2$ need to be extremely sparse, since $H_2 H_1$ is typically much denser than any of the two. Let us illustrate this observation on a simple example. 
\begin{example}[The product of matrices satisfying the SSC typically does not satisfy the SSC] 
Let $r_2 = 3$, $r_1 = 6$, and 
\[ 
H_2 = 
\left( \begin{array}{cccccc}
\omega & 1 & 1 & \omega & 0 & 0 \\ 
1 & \omega & 0 & 0 & \omega & 1 \\ 
0 & 0 & \omega & 1 & 1 & \omega \\ 
\end{array}
\right). 
\]
It can be shown that $H_2$ satisfies the SSC if and only if $\omega < 0.5$~\cite{huang2013non}. 
The matrix $H_2$ is the sparsest non-trivial case for a rank-three matrix that satisfies the SSC, since having columns with two zero entries \revises{corresponds to} a stronger condition, referred to as separability, which makes NMF much easier to handle, because columns of $W$ are present in the data set, up to scaling~\cite{arora2012computing}.   
Now, for any matrix $H_1 \in \mathbb{R}^{6 \times n}$ having 3 nonzeros entries per column, $H_2 H_1$ \revises{will not have any entry equal to zero}, for any $\omega > 0$ (since the sum of any 3 columns of $H_2$ is dense). Hence $H_2 H_1$ cannot satisfy the SSC for any matrix $H_1$ with 3-sparse columns\revises{, that is, 3 non-zero entries per columns, as the SSC requires some degree of sparsity (see above)}, and hence the factorization $X = W_2 (H_2 H_1)$ will not be unique. In fact, a necessary condition for the uniqueness of the NMF $X = WH$ is that the supports of the rows of $H$ are not contained in one another~\cite{huang2013non}. 

On the other hand, matrices $H_1$ with three non-zeros per column, and sufficiently many columns, are likely to satisfy the SSC. For example, we generated randomly 1000 matrices with 100 columns and with 3 non-zeros entries per column, where the position of the non-zero entries are randomly selected, and the non-zero entries are generated with the uniform distribution in the interval [0,1]. Among these 1000 sparse matrices, all satisfied the SSC\footnote{Although it is NP-hard to check the SSC~\cite{huang2013non}, it is possible to do it for medium-scale matrices by solving a non-convex quadratic optimization problem with Gurobi. We thank Robert Luce, from Gurobi, to help us write down and solve this optimization problem.}. 
\end{example}
In summary, the consideration of the product of the $H_\ell$ matrices within the factorizations in DCLF makes it less likely to achieve identifiability compared to LCLF. This is primarily due to the tendency of these products to become increasingly denser as the factorization unfolds. 


\section{Deep $\beta$-NMF: models and algorithms}
\label{sec:algorithm} 
In this section, we propose two new deep $\beta$-NMF models, describe the algorithms for solving them, and consider the convergence guarantee for the algorithms. The $\beta$-divergence between two matrices $A$ and $B$ is defined as
\begin{equation*}
    D_\beta(A,B) = \sum_{i,j} d_\beta(A_{ij},B_{ij}), 
\end{equation*}
where, for scalars $x$ and $y$, 
\begin{equation} \label{eq:betadivergence}  
d_{\beta}(x,y) = \left\{ 
\begin{array}{cc}
  \frac{x}{y} - \log \frac{x}{y} - 1  & \text{ for }  \beta = 0, \\
 x \log \frac{x}{y} - x + y & \text{ for } \beta = 1, \\ 
\frac{1}{\beta (\beta-1)} \left(x^\beta + (\beta-1) y^\beta - \beta x y^{\beta-1}\right) &  \text{ for } \beta \neq 0,1,  
 \end{array}
\right. 
\end{equation} 
\revise{see \cite{basu1998robust, eguchi2001robustifying}}. 
When $\beta = 2$, this corresponds to the least-squares measurement, whereas for $\beta = 1$, it corresponds to the Kullback-Leibler (KL) divergence. With convention that $a\times\log 0=-\infty$ for $a>0$ and $0 \times \log 0=0$, the KL-divergence is well-defined. 

\subsection{The two proposed deep NMF models}

As elucidated in Section~\ref{sec:ident_deepNMF}, it is more \revise{likely} for LCLF to be identifiable \revise{compared} to   DCLF. Hence we employ LCLF as the basis, and  introduce the following two novel deep $\beta$-NMF model: 
\begin{enumerate}
    \item Deep $\beta$-NMF without regularization:
    \begin{equation} \label{eq:probBetaNMF}
\min_{\{W_\ell \geq 0, H_\ell \geq 0\}_{\ell=1}^L}
\sum_{\ell=1}^L \lambda_\ell \revise{D_\beta}(W_{\ell-1}, W_\ell H_\ell)
\; \text{ subject to }  \; 
H_{\ell} \ e = e \text{ for } \ell=1,2,\dots,L, 
\end{equation} 
where $W_0 = X$, $W_\ell$ has $r_\ell$ columns, $r_0 = n$, and the $\lambda_\ell$'s are positive penalty parameters, and $e$ is the vector of all ones of appropriate dimension. \\

\textbf{Why the normalization $H_{\ell} \ e = e$?}
Let us explain why we choose the normalization constraints $H_{\ell} \ e = e$ in our deep $\beta$-NMF model without regularization. The LCLF~\eqref{eq:lossLCLF} is not fully consistent because of the scaling degree of freedom in NMF, this was not pointed out in~\cite{de2023consistent}. In fact, except at the first layer, all errors $\revise{D_\beta}(W_{\ell-1}, W_\ell H_\ell)$ for $\ell=2,3,\dots,L$ can be made arbitrarily small by using the scaling degree of freedom: multiply $W_1$ by an arbitrarily small positive constant and divide $H_1$ by the same constant. This does not change $\revise{D_\beta}(X, W_1 H_1)$ while $W_1$ is arbitrarily close to zero making $\revise{D_\beta}(W_1, W_2 H_2)$ arbitrarily small (for any norm which is not scaled invariant, e.g., all $\beta$-divergences for $\beta > 0$).  Therefore, for~\eqref{eq:lossDCLF} to make sense, it is crucial to add a normalization constraints on the $W_\ell$'s or the $H_\ell$'s. 
Many options are possible, and depend on the application at hand. For example, in hyperspectral imaging, we might impose $H_\ell^\top e = e$ for all $\ell$ which is known as the sum-to-one constraint of the abundances, 
and, in topic modeling, where the columns of $W_\ell$ correspond to topics,  
we might impose 
 $W_\ell^\top e = e$ as the entries in each  column of $W_\ell$ correspond to probabilities of the words to belong to the corresponding topic. 
In this paper, we choose $H_\ell \ e = e$, that is, the sum of the entries in each row of $H_\ell$ sums to one, as in~\cite{fu2018identifiability}. 
The main reason is that this normalization can be made w.l.o.g.\ by the scaling degree of freedom. Moreover, we do not constraint $W_\ell$ as it would make the design of closed-form MU for $W_\ell$ much more difficult, if possible (see Section~\ref{sec:algominvol} for such a case).

\item Minimum-volume (min-vol) deep $\beta$-NMF: \begin{equation} \label{eq:ourmodel}
\min_{\revises{\{W_\ell \geq 0, H_\ell \geq 0\}_{\ell=1}^L}}
\sum_{\ell=1}^L \lambda_\ell D_\beta(W_{\ell-1}, W_\ell H_\ell) + \alpha_\ell 
\log\det\left( W_{\ell}^\top W_{\ell} + \delta I \right) 
\; \text{ s.t. }  \; 
W_{\ell}^\top e = e, 
\end{equation} 
where $W_0 = X$, $W_\ell$ has $r_\ell$ columns with $r_0 = n$, $\delta$ is small positive scalar that prevents the $\log\det$ to go to $-\infty$, and the $\alpha_\ell$'s are positive penalty parameters. 
The choice to normalize the columns of $W_\ell$, rather than the rows of $H_\ell$, is from the fact that it significantly improves the conditioning of the min-vol NMF problem. When rows of $H_\ell$ are normalized, it can lead to highly ill-conditioned of $W_\ell$, especially when certain columns of $W_\ell$ exhibit substantially larger norms compared to others. This issue is further elaborated 
 in~\cite{leplat2020blind} and~\cite[Chapter 4.3, 3.5]{gillis2020nonnegative}. 
 \end{enumerate}


 \subsection{Algorithms for solving the proposed deep $\beta$-NMF models}
 
Obtaining a global solution for deep NMF is a computationally challenging problem, as it generalizes NMF which is NP-hard~\cite{Vavasis2010complexity}. 
Moreover, the objective functions presented in \eqref{eq:probBetaNMF} and \eqref{eq:ourmodel} are jointly non-convex for the variable $(W_1, \dots, W_L, H_1, \dots, H_L)$. Consequently, updating all factors simultaneously can be prohibitively expensive. Therefore, most algorithms designed to address  (deep) NMF rely on block coordinate methods. These methods update one factor at a time while keeping the others fixed. In this paper, we also adopt this strategy to efficiently address the optimization challenges associated with deep NMF. Specifically, we will employ the block majorization minimization (BMM) framework, which was designed to solve the following multi-block nonconvex optimization problem: 
\begin{equation} 
\label{eq:composite}
\min_{ x := (x_1,x_2,\dots,x_s) \in \mcalX} F(x):= f(x) + \sum_{i=1}^s g_i(x_i), 
\end{equation}  
where $f$ is continuous on $\mathcal X$, $g_i$'s are  proper and lower-semicontinuous functions (possibly with extended values), and  $\mathcal X = \mathcal X_1 \times \mathcal X_2 \times \dots \times \mathcal X_s$ with $\mathcal X_i$ ($i=1,2,\dots,s$) being closed convex sets. At iteration $k$, BMM fixes the latest values of block $j\ne i$ and updates block $x_i$  by  
\begin{equation}
\label{eq:MM}
x_i^{k} \in \arg\min_{x_i\in \mathcal X_i} \left \{u_i(x_i,x_1^k,\ldots,x_{i-1}^k,x_i^{k-1}, x_{i+1}^{k-1},\ldots,x_s^{k-1}) + g_i(x_i)\right \},
\end{equation}
where $u_i : \mathcal X_i \times \mathcal X \to \mathbb R$ is a block majorizer of $f(x)$, that is, $u_i$ satisfies the following conditions    
\begin{equation}
\label{eq:majorize}
\begin{split}
&u_i (x_i , x) = f(x), \forall \, x \in \mathcal X, 
\\ 
&u_i (y_i , x) \geq f(x_1,\ldots,x_{i-1},y_i,x_{i+1},\ldots,x_s), \forall \, y_i \in \mathcal{X}_i,  x \in \mathcal X.
\end{split}
\end{equation}
From the definition of $u_i$, we have
\begin{align*}
\begin{split}
F(x^k)&=u_1(x_1^k,x_1^k,\ldots,x_s^k) + \sum_{i} g_i(x_i^k) \geq u_1(x_1^{k+1},x_1^k,\ldots,x_s^k) +g_1(x_1^{k+1})+\sum_{i\ne 1} g_i(x_i^k)\\&\geq f(x_1^{k+1},x_2^k,\ldots,x_s^k) +g_1(x_1^{k+1})+\sum_{i\ne 1} g_i(x_i^k)\\
&=u_2(x_2^k,x_1^{k+1},x_2^k,\ldots,x_s^k)+g_1(x_1^{k+1})+\sum_{i\ne 1} g_i(x_i^k)\\
&\geq u_2(x_2^{k+1},x_1^{k+1},x_2^k,\ldots,x_s^k) + g_1(x_1^{k+1}) + g_2(x_2^{k+1}) + \sum_{i\not\in \{1,2\}} g_i(x_i^k)\\
&\geq f(x_1^{k+1},x_2^{k+1},\ldots,x_{s-1}^k,x_s^k) + g_1(x_1^{k+1}) + g_2(x_2^{k+1}) + \sum_{i\not\in \{1,2\}} g_i(x_i^k)\\
&\geq  \ldots\geq F(x^{k+1}).
\end{split}
\end{align*}
In other words, BMM produces a non-increasing sequence $\{F(x^k)\}$.
We refer the readers to \cite{HPG_TITAN,Mairal_ICML13,RHL_BSUM,Sun2017_TSP} for examples of majorizer functions. BMM was introduced in \cite{RHL_BSUM} (with the name BSUM - block successive upper-bound minimization) to solve nonconvex problem \eqref{eq:composite} with $g_i=0$ for $i=1,\ldots,s$. It is proved in \cite[Theorem 2]{RHL_BSUM} that if the following conditions are satisfied then we have a convergence guarantee for BSUM: 
\begin{itemize}
    \item $u_i(y_i,x)$ is quasi-convex in $y_i$ for $i=1,\ldots,s$,
    \item the subproblem \eqref{eq:MM} with $g_i=0$ has a unique solution,
    \item $u_i'(y_i,x;d_i)\big |_{y_i=x_i}=f'(x;d)\quad \forall d=(0,\ldots,d_i,\ldots,0)$ s.t. $x_i + d_i \in \mcalX_i$ $\forall i$,
    \item $u_i(y_i,x)$ is continuous in $(y_i,x)$, for all $i$.
\end{itemize}

\revises{It is worth mentioning that TITAN introduced in \cite{HPG_TITAN}  and BMMe introduced in \cite{hien2024block} are 
 accelerated versions of BMM for solving Problem \eqref{eq:composite}. TITAN and BMMe enhance BMM by employing inertial terms/extrapolations in each block update, which significantly boosts the convergence of BMM.}  Leveraging the convergence outcomes established for BSUM, \revises{TITAN, and BMMe}, numerous algorithms addressing low-rank factorization problems come with guaranteed convergence. For example, BSUM assures the convergence of a perturbed Multiplicative Update (MU) and a block mirror descent method for KL NMF, see~\cite{HienNicolas_KLNMF}; TITAN provides convergence guarantees for accelerated algorithms dealing with min-vol NMF~\cite{Vu-Thanh2021}, sparse NMF and matrix completion~\cite{HPG_TITAN}; \revises{BMMe guarantees convergence of MU with extrapolation for $\beta$-NMF with $\beta\in [1,2]$.} 

Although convergence guarantees have been firmly established for BSUM (\revises{TITAN and BMMe}) under appropriate assumptions, which serve as valuable tools to ensure the convergence of BMM in solving deep NMF models, it is not a straightforward task to construct suitable majorizers that satisfy the required assumptions. In the next sections, we propose suitable majorizers and apply BSUM to design efficient algorithms to solve the two proposed deep $\beta$-NMF models. To that end, we need the following lemma from \cite{fevotte2011algorithms} that provides a majorizer for $h\mapsto \sum_i d_\beta(v_i , [Wh]_i)$, where vector $v$ and matrix $W$ are fixed. 

\begin{lemma}[Majorizer function for $\beta$-NMF~\cite{fevotte2011algorithms}] \label{defG} 
Denote $W \tilde h$ by $\tilde v$ and the entries $[W \tilde h]_i$ by $\tilde v_i$. Let $\tilde h $ be such that $\tilde v_i> 0$ and $\tilde h_i>0$. Then the following function 
\begin{equation}\label{eq:20}
\begin{aligned}
G(h,\tilde h)&=\sum_i\Big[ \sum_j \frac{w_{ij} \tilde h_j}{\tilde v_i} \check{d} \big(v_i, \tilde v_i \frac{h_j}{\tilde h_j} \big)\Big] \\ 
& \qquad + \Big[\hat{d}^{'}(v_i, \tilde v_i)\sum_j w_{ij} (h_j -\tilde h_j) + \hat{d}(v_i, \tilde v_i)\Big] + \bar{d}(v_i)
\end{aligned}
\end{equation} 
is a majorizer of the function  $h\mapsto \sum_i d_\beta(v_i , [Wh]_i)$, 
 where $\check{d}$ is a convex function of $u$, $\hat{d}$ is a concave function of $u$ and $\bar{d}$ is a constant of $u$ in the following decomposition of $d_\beta$  
\begin{equation}\label{eq:3}
  d_{\beta}(v,u)= \check{d}(v,u)+\hat{d}(v,u)+\bar{d}(v), 
  \end{equation}
see Table~\ref{table:conv_concav_decomp}.
\begin{center}
\begin{table}[h!]
\begin{center}
\caption{Differentiable convex-concave-constant decomposition of the $\beta$-divergence under the form \eqref{eq:3}~\cite{fevotte2011algorithms}. }
\label{table:conv_concav_decomp}
\begin{tabular}{|c|c|c|c|}
\hline 
      & $\check{d}(v,u)$  &  $\hat{d}(v,u)$ & $\bar{d}(v)$   \\  \hline 
      \revises{$\beta=(-\infty,1) \setminus \{0\}$ }     
 &   \revises{$\frac{1}{1 - \beta} v u^{\beta-1}$}   & \revises{$\frac{1}{\beta} u^{\beta}$}  & \revises{$\frac{1}{\beta (\beta-1)} v^\beta$}  \\
 $\beta=0$      
 & $vu^{-1}$     & $\log(u)$  & $u(\log(v)-1)$  \\
 $\beta \in [1,2] $ 
 & $d_{\beta}(v,u)$ & 0 & 0    \\ \hline  
\end{tabular} 
\end{center}
\end{table}
\end{center}
\end{lemma}
\begin{note}
\label{note1}
It is important to note that $G(h,\tilde h)$ is convex in $h$. Furthermore, since we have
${D_\beta(y,wH)=D_\beta(y^\top,H^\top w^\top)}$, where vector $y^{\top}$ and matrix $H$ are fixed, Lemma~\ref{defG} can be used to derive a similar majorizer $G(w,\tilde w)$ for $w\mapsto  D_\beta(y,wH)$.  On the other hand, note that $ D_\beta(Y,WH)=\sum_i D_\beta (y_i,w_iH)$, where $y_i$ and $w_i$ are the $i$-th row of $Y$ and $W$ respectively. This means $W\mapsto D_\beta(Y,WH)$ is separable with respect to the rows of $W$. Hence, we can formulate a majorizer for $W\mapsto D_\beta(Y,WH)$ by summing up the majorizers of its rows $w\mapsto  D_\beta(y,wH)$. We have similar procedure for $H\mapsto D_\beta(Y,WH)$.  Considering KL NMF, BSUM using the majorizers in Lemma~\ref{defG} is the MU algorithm proposed in \cite{Lee1999Learning,lee2001algorithms}, see \cite{HienNicolas_KLNMF}. 
\end{note}



\subsubsection{Algorithm for solving deep $\beta$-NMF without regularization} \label{sec:algoNoregul}


Problem~\eqref{eq:probBetaNMF} has the form of \eqref{eq:composite}, where $x$ comprises $W_l$ and  $H_l$ for $l=1,\ldots,L$,  $g_i=0$, \revise{and} the closed convex set $\mcalX_i$ \revise{corresponding} to $W_l$ is $\mathbb R^{m\times r_l}_+$ and that \revise{corresponding} to $H_l$ is $\{H_l: H_l\in \mathbb R^{r_l \times r_{l-1}}_+, H_l e = e\}$.

\paragraph{Update of $H_l$} We observe that $H_l$ only appears in one term of the objective function in~\eqref{eq:probBetaNMF}. While fixing the other factors, minimizing the objective of Problem~\eqref{eq:probBetaNMF} with respect to $H_l$ is the same as in standard $\beta$-NMF. We hence employ the majorizers in Lemma~\ref{defG} (see also Note~\ref{note1}) and the recently introduced framework in~\cite{Leplat2021} that allows one to derive MU for block of variables satisfying disjoint equality constraints as well as nonnegativity constraints.  See Algorithm~\ref{ADKL} for the actual update. Note that, in the update of $H_\ell$, the parameter $\mu$ appearing in the denominator correspond to the optimal vector of Lagrange multipliers allowing the new updates to satisfy both the nonnegativity and the sum-to-one constraints, see \cite{Leplat2021} for more details about the procedure for such updates.

\paragraph{Update of $W_l$ for $l=1,\ldots,L-1$.} 
While fixing the other factors, the corresponding subproblems of \eqref{eq:probBetaNMF} with respect to each of the block $W_\ell$, $l=1,\dots,L-1$, have the same structure. Hence we can focus on building the majorizer for one representative $W_\ell$. To simplify the notation, let us denote $W_\ell$ by $W$, $W_{\ell-1}$ by $Y$, $H_\ell$ by $H$, and $W_{\ell+1} H_{\ell+1}$ by $\Bar{W}$.  
Then each subproblem is equivalent to the following problem (after removing the constants in the objective): 
\begin{equation}\label{eq:probinWl}
\text{find}  \quad \underset{W \geq 0}{\argmin}  \; \lambda_l D_{\beta}(Y,W H) 
  + \lambda_{l+1} D_{\beta}(W,\bar{W}), 
\end{equation} 
where $Y$,  $H$, $\bar{W}$ are given and kept fixed during the update of $W$. 
Let $w_i$, $y_i$, and $\bar w_i$, $i=1,\ldots,m$, be the rows of $W$, $Y$ and $\bar W$ respectively. Note that the objective function of  \eqref{eq:probinWl} equals to 
$$\lambda_l \sum_i^m  D_{\beta}(y_i , w_i H) + \lambda_{l+1} \sum_i^m D_{\beta}(w_i,\Bar{w}_i),$$
which is separable with respect to the rows of $W$. 
Therefore, we can focus on building the majorizer for each row of $W$ (then sum up these majorizers to formulate the majorizer for $W$). 
Problem~\eqref{eq:probinWl} restricted to a particular row $i$ of $W$ is equivalent to the following problem: 
\begin{equation}\label{eq:probinw}
 \text{find}\quad \underset{w \geq 0}{\argmin}\;  \lambda_\ell  D_{\beta}(y,w H) + \lambda_{\ell+1}  D_{\beta}(w,\bar{w}), 
\end{equation}
where the subscript $i$ has been dropped for notation succinctness. 
For the first term in the objective of \eqref{eq:probinw}, we use the majorizer $G(w,\tilde w)$ proposed in \cite{fevotte2011algorithms},  see Note~\ref{note1}. 
This implies that $\lambda_\ell G(w,\tilde w) + \lambda_{\ell+1} D_\beta (w,\bar w)$ is a majorizer of $w\mapsto \lambda_\ell D_\beta(y, wH) + \lambda_{\ell+1} D_\beta (w,\bar w)$. Consequently, the update of each row of $W$ is  
\begin{equation}\label{eq:probinhupper}
  w\in\underset{w \geq 0}{\argmin} \quad  G(w,\tilde{w}) + \lambda D_{\beta}(w,\Bar{w}),
\end{equation}
where $\lambda=\frac{\lambda_{\ell+1}}{\lambda_\ell}$.  
To compute the positive minimizer of \eqref{eq:probinhupper}, it's sufficient to look for $w \in \mathbb{R}^{r_\ell}_+$ that cancels the gradient of objective function from \eqref{eq:probinhupper}. Since the objective function is separable w.r.t. each entry $w_{k_\ell}$ of $w$, we focus on solving:
\begin{equation}\label{eq:gradTocancel}
    \text{find } \hat{w}_{k_\ell}\geq 0 \text{ such that } \nabla_{w_{k_\ell}}\left[  G(w,\tilde{w}) + \lambda D_{\beta}(w,\Bar{w}) \right]_{w=\hat{w}_{k_\ell}} = 0. 
\end{equation}

The next steps depend on the value chosen for $\beta$. A closed-form of the minimizer $\hat{w}$ can be derived for  $\beta \in \{0,1/2,1,3/2,2\}$. 
 Note that considering the case $\beta=2$ is excluded since the objective function is $L$-smooth in this case, hence efficient first-order methods can be used to tackle the subproblems such as the well-known Nesterov accelerated projected gradient descent~\cite{Nesterov2018}. Furthermore, for $\beta\in\{0,1/2,1,3/2\}$, $D_\beta(w,\tilde w)$ is strictly convex in $w$, which makes the majorizer $\lambda_\ell G(w,\tilde w) + \lambda_{\ell+1} D_\beta (w,\bar w)$ strictly convex in $w$ (this convexity is used to verify the conditions for a convergence guarantee of BSUM).   In the following, we detail the updates for the case $\beta = 1$. The updates  for $\beta\in \{0,\revises{1/2},3/2\}$ can be found in Appendix~\ref{sec:append}.    
 
 Considering $\beta=1$, from Lemma \ref{defG} and Table~\ref{table:conv_concav_decomp} for $\beta =1$, one can show that solving \eqref{eq:gradTocancel} boils down to find a nonnegative solution $w_{k_\ell}$ of the following scalar equation:
\begin{equation}\label{eq:updateh_kl}
    a = \frac{b}{w_{k_\ell}} - \lambda \log(w_{k_\ell}),     
\end{equation}
where $a=\sum_{j_\ell} H_{k_{\ell}j_\ell} - \lambda \log(\Bar{w}_{k_\ell})$  
and $b=\tilde{w}_{k_\ell} \sum_{j_\ell} H_{k_{\ell}j_\ell} \frac{y_{j_\ell}}{[\tilde{w}H]_{j_\ell}}$(note that $y_{j_\ell}$ is an entry of a row $y$).
Equation \eqref{eq:updateh_kl} has the following  nonnegative solution:
\begin{equation}\label{eq:sol_updateh_kl}
    \hat{w}_{k_\ell} = \frac{b}{\lambda \mathcal{W} \left( \frac{b e^{\frac{a}{\lambda}}}{\lambda}\right)}, 
\end{equation}
where $\mathcal{W}(.)$ denotes the Lambert $\mathcal{W}$-Function. \revise{Note that it is typical to use the Lambert function to describe the solution of \eqref{eq:updateh_kl}; see  \cite{Corless1996}.} Interestingly, this update is well defined at the boundaries of the feasible set, in particular when $b$ and $a$ respectively tend to 0 and $+\infty$, the latter occurs when the entry $\Bar{w}_{k_\ell}$ tends to 0. Indeed,  we have $\lim\limits_{\substack{b \to 0, b>0 \\ a<+\infty}} \hat{w}_{k_\ell}(b)=e^{-\frac{a}{\lambda}}$, and $\lim\limits_{\substack{a \to +\infty \\ b\neq 0}} \hat{w}_{k_\ell}(a)=0$.
Equation \eqref{eq:sol_updateh_kl} can be expressed in matrix form as follows:  
\begin{equation}\label{eq:updateW}
    \begin{aligned}
       \hat{W}=\frac{\left[B \right]}{\left[\lambda \mathcal{W} \left( \frac{B \odot e^{.\frac{A}{\lambda}}}{\lambda}\right)\right]}, 
    \end{aligned}
\end{equation} 
where $A=J H^\top_\ell  - \log(\Bar{W})$ with $J$ is a all-one matrix of size $m$-by-$r_{\ell-1}$, and $\log$ is element wise, and the notation $e^{.A}$ is the element-wise exponential, and $B=\tilde{W} \odot \left( \frac{\left[ Y \right]}{\left[ \tilde{W}H \right]}H^\top \right) $ with $C \odot D$ (resp. $\frac{\left[ C \right]}{\left[ D \right]}$ ) is the Hadamard product (resp. division) between $C$ and $D$ and $C^{(.\alpha)}$ is the element-wise $\alpha$ exponent of $C$. It is important to note that the MU update in \eqref{eq:updateW} would encounter a zero locking phenomenon, that is, it is unable to make changes to an entry within $\tilde{W}$ when that entry equals 0 (since when $ \tilde w_{k_\ell}=0$ we would have $b=0$). This issue can be resolved by selecting an initial $W$ that contains strictly positive entries.



\paragraph{Update of $W_L$} We observe that $W_L$ only appears in the last term of the objective in \eqref{eq:probBetaNMF}.  Minimizing the objective of Problem~\eqref{eq:probBetaNMF} with respect to $W_L$ is the same as in standard $\beta$-NMF.  As noted in Note~\ref{note1}, BSUM step collapses to the classical MU of $\beta$-NMF. 

Algorithm~\ref{ADKL} summarizes our proposed algorithm for deep $\beta$-NMF in the case $\beta = 1$, that is, for deep KL-NMF, recalling that $W_0 = X$ and $W_\ell$ has $r_\ell$ columns with $r_0 = n$. 

\renewcommand{\baselinestretch}{1}
\algsetup{indent=2em}
\begin{algorithm}[ht!]
\caption{Algorithm for Deep KL-NMF \label{ADKL}}
\begin{algorithmic}[1] 
\REQUIRE Input matrix $X$, number of layers $L$, inner ranks $r_l$'s, weight parameters $\lambda_{\ell}$'s.
\ENSURE An approximate solution to Problem \ref{eq:probBetaNMF}
    \medskip  
\FOR{$k=1,...$}
    \FOR{$l\in\{1,...,L\}$} 
        \STATE \emph{\% Update of factors $H_\ell$ using \cite{Leplat2021}}
        \STATE $H^k_{\ell} \leftarrow H^{k-1}_{\ell} \odot \frac{
        \left[W_\ell^{T,k-1} \left( \frac{\left[W^{k}_{\ell-1}\right]}{\left[W^{k-1}_{\ell}H^{k-1}_{\ell}\right]}\right)\right]
        }{
        \left[W_{\ell}^{T,k-1} e_{m \times r_{\ell-1}} - \mu e_{1 \times r_{\ell-1}}\right] 
        }, $
        where $\mu$ is the root of a univariate polynomial. 
        \STATE \emph{\% Update of factors $W_\ell$}
        \IF{$l<L$}
            \STATE $\lambda \leftarrow \frac{\lambda_{\ell+1}}{\lambda_{\ell}}$
            \STATE $B \leftarrow W^{k-1}_{\ell} \odot \left( \frac{\left[ W_{\ell-1}^k \right]}{\left[ W^{k-1}_{\ell}H_\ell^k \right]}H_\ell^{T,k} \right) $
            \STATE $A \leftarrow J H^{T,k}_\ell  - \log(W_{\ell+1}^{k-1}H_{\ell+1}^{k-1})$ 
            \STATE $W^k_{\ell} \leftarrow \frac{\left[B \right]}{\left[\lambda \mathcal{W} \left( \frac{B \odot e^{.\frac{A}{\lambda}}}{\lambda}\right)\right]}$
            
        \ELSIF{$l=L$}
            \STATE $W^k_{L} \leftarrow W^{k-1}_{L} \odot \frac{\left[ \left( \frac{[W^{k}_{L-1}]}{[W^{k-1}_{L}H^{k}_{L}]}\right)H_L^{T,k}\right]}{\left[e_{m \times r_{L}} H_{L}^{T,k} \right]}$, which are the standard MU for KL-NMF. 
        \ENDIF
        
    \ENDFOR
\ENDFOR
\RETURN $\{W_L,H_L,H_{L-1},...,H_1\}$
\end{algorithmic}  
\end{algorithm} 
\renewcommand{\baselinestretch}{2}

\paragraph{Convergence guarantee} Although Algorithm~\ref{ADKL} produces a non-increasing sequence $\{f^k\}$, where $f^k$ is the objective of \eqref{eq:probBetaNMF} at iteration $k$ (since its update follows BSUM), it is not guaranteed that the generated sequence of Algorithm~\ref{ADKL} converges (the objective in \eqref{eq:probBetaNMF} is not directionally  differentiable). To have some convergence for Algorithm~\ref{ADKL}, we need to impose the constraints $W_\ell \geq \varepsilon$ and $H_\ell \geq \varepsilon$, where $\varepsilon>0$, to \eqref{eq:probBetaNMF}. In our implementation, we choose the MATLAB machine epsilon for $\varepsilon$. Using the same majorizers, we obtain a perturbed version of Algorithm~\ref{ADKL}, in which we take the element-wise maximum between the updates of factors $\{W_\ell,H_\ell\}_{\ell=1}^L$ (corresponding to the closed-form expression of the minimizer of BMM step built at the current iterate) and $\varepsilon$. With this additional constraints, the sufficient conditions for convergence of BSUM would be satisfied, leading to the convergence of the perturbed version of Algorithm~\ref{ADKL}.

Note that a similar rationale can be applied to have a convergence guarantee for a perturbed version of the algorithm solving the proposed deep $\beta$-NMF with $\beta\in\{0,3/2\}$.
In that case, we also have closed-form solution for \eqref{eq:gradTocancel} and strict convexity for the majorizer used in \eqref{eq:probinhupper}. Hence, a similar rationale can be applied to have a convergence guarantee for a perturbed version of the algorithm solving the proposed deep $\beta$-NMF, with $\beta\in\{0,3/2\}$. For the sake of completeness, we provide in Appendix~\ref{sec:append} the updates for $\{W_l\}_{l=1}^{L}$ factors when $\beta\in\{0,3/2\}$.



\paragraph{Parallelization} The proposed algorithm relies on MU-based approaches, which involve computationally intensive steps in matrix products. Furthermore, the updates to all factor entries can be carried out independently. Therefore, the proposed algorithm can be effectively executed on a parallel and high performance computing platform.

\paragraph{Usefulness of our MU in other contexts}  
    The MU~\eqref{eq:updateW} allows us to update $W$ to minimize~\eqref{eq:probinWl}. 
    The regularized NMF problem~\eqref{eq:probinWl}, in which $\bar{W}$ is a given matrix to which $W$ should be close to,  could be useful in other contexts than deep $\beta$-NMF, e.g., 
    if some $\bar{W}$ is available via prior some knowledge, 
    by taking $\bar{W} = 0$ to regularize $W$, 
    or 
    for symmetric NMF where $\bar{W} = H^\top$~\cite{Zhihui21}.  
    A very similar regularization has appeared in temporal NMF, see~\cite{fevotte2018temporal}, where columns of $H$ correspond to different times and hence are highly correlated, therefore using  $\sum_{j=1}^{n-1} D_\beta\big( H(:,j),H(:,j+1) \big)$ as a regularization allows one to account for temporal dependencies. 
Another occurrence of such models is co-factorizations, where two or more matrices are factorized with shared factors; see~\cite{seichepine2014soft, gouvert2018matrix}.


\subsubsection{Algorithm for solving minimum-volume deep \revise{KL-NMF} } 
\label{sec:algominvol}


Problem~\eqref{eq:ourmodel} has the form of \eqref{eq:composite} in which $x$ comprises $W_l$ and  $H_l$ for $l=1,\ldots,L$,  $g_i=0$, the closed convex set $\mcalX_i$ that  corresponds to $W_\ell$ is $\{W_\ell: W_\ell\in \mathbb R^{m \times r_{\ell-1}}_+, W_\ell^\top e = e\}$  and that  corresponds to  $H_\ell$ is $\mathbb R^{r_{\ell}\times r_{\ell-1}}_+$. \revise{In the following, } let us focus on the special case $\beta=1$. Similar rationale can be followed for other values of $\beta$. 

\paragraph{Update of $H_\ell$}
Each factor $H_{\ell}$ with $1\leq \ell \leq L$ only appears in a single term within the objective function of Problem \eqref{eq:ourmodel}. As a result, one can directly apply the classical MU updates for $H_{\ell}$.

\paragraph{Update of $W_\ell$, $\ell=1,\ldots,L-1$}
Since all the subproblems in $W_{\ell}$ with $1\leq\ell<L$ have the same structure, we detail in the following the methodology for deriving the BSUM update for one specific $W_{\ell}$. 
 For notation succinctness, we drop the subscript $\ell$ and denote $W_\ell$ by $W$, $W_{\ell-1}$ by $Y$, $H_\ell$ by $H$, and $W_{\ell+1} H_{\ell+1}$ by $\Bar{W}$. The subproblem in $W$ is as follows:
\begin{equation*}
    \begin{aligned}
\text{ find}\quad      \arg\min_{W}
& \quad \lambda_\ell \revise{D_{KL}}(Y, W H) + \lambda_{\ell+1}  \revise{D_{KL}}(W, \bar{W}) + 
\alpha_\ell
\log\det\left( W^\top W + \delta I \right) \\
                                      &\text{such that } W^\top e = e, W \geq 0.
    \end{aligned}
\end{equation*}
For a given matrix $\tilde{W}$, denote $A = (\tilde{W}^{\top}\tilde{W} + \delta I)^{-1}$,  $A^+=\max(A,0) \geq 0$, $A^-=\max(-A,0) \geq 0$, and $\Phi(\tilde \omega_i)=\text{Diag}\left(2\frac{[A^+ \tilde w_i + A^- \tilde w_i]}{[\tilde w]}\right)$ (which is a diogonal matrix), where $\tilde w_i$ is a row of $\tilde W$, and $\frac{[\cdot]}{[\cdot]}$ denotes the component-wise
division. Let $l(w) =w^\top A w$ and $\Delta w_i=w_i-\tilde w_i$, where $w_i$ is a row of $W$. From \cite[Lemma 3]{leplat2020blind}, we have  
\begin{equation}
    \label{logdet_majorizer}
    \begin{aligned}
&\log\det\left( W^\top W + \delta I \right) \leq\log\det\left( \tilde{W}^\top \tilde W + \delta I \right) + \langle A, W^\top W -\tilde{W}^\top \tilde W  \rangle\\
&\leq g(W,\tilde W):=\log\det\left( \tilde{W}^\top \tilde W + \delta I \right) -\langle A, \tilde{W}^\top \tilde W  \rangle + \sum_{i} l(\tilde w_i)+\langle\Delta w_i,\nabla l(\tilde w_i)\rangle\\
&\qquad\qquad\qquad\qquad\qquad+\frac12\langle \Delta w_i,\Phi(\tilde w_i)\Delta w_i \rangle.
\end{aligned}
\end{equation}
We then use the following majorizer for $W$:
$$\bar G(W,\tilde W)= \lambda_\ell G(W,\tilde W) + \lambda_{\ell+1}  \revise{D_{KL}}(W, \bar{W}) + \alpha_l g(W,\tilde W),
$$
where $G(W,\tilde W)$ is the majorizer of $W\mapsto \revise{D_{KL}}(Y,WH)$, which is formed by summing the majorizers proposed in Lemma~\ref{defG} for each row of $W$ as discussed in Note~\ref{note1}. As $G(W,\tilde W)$ and $\revise{D_{KL}}(W, \bar{W})$ are convex  in $W$ and $g(W,\tilde W)$ is strongly convex in $W$  over $\mathbb R_+^{{m} \times r_{\ell-1}}$, we also have $G(W,\tilde W)$ is strongly convex in $W$. 
Hence, the following subproblem 
 for the update of $W$ has the unique solution: 
\begin{equation}\label{eq:subProblem_minvolWl}
    \begin{aligned}
 \text{find}    \quad \arg\min_{W}
& \quad \lambda_\ell G(W, \tilde W) + \lambda_{\ell+1}  \revise{D_{KL}}(W, \bar{W}) + 
\alpha_\ell
 g(W,\tilde W) \\
                                      &\text{such that } W^\top e = e, W \geq 0,
    \end{aligned}
\end{equation}
where $\tilde W$ is the current iterate. Below we will describe an ADMM to solve Problem \eqref{eq:subProblem_minvolWl}.  
\paragraph{Update of $W_L$} We observe that $W_L$ only appears in the last term of the objective. We use the following majorizer for $W_L$
$$\bar G_L(W_L,\tilde W_L)=\lambda_L  G_L(W_L, \tilde W_L) + \alpha_L g(W_L,\tilde W_L), 
$$
where $G_L(W_L,\tilde W_L)$ is the majorizer of $W_L\mapsto \revise{D_{KL}}(W_{L-1},W_L H_L)$, which is derived as in Note \ref{note1}, and $g$ is defined in \eqref{logdet_majorizer}. \revise{Then the update of $W_L$ is the MU proposed in~\cite{Leplat2021}}.

\paragraph{ADMM for solving Problem \eqref{eq:subProblem_minvolWl} to update $W_\ell$} We rewrite   Problem \eqref{eq:subProblem_minvolWl} as follows: 
\begin{equation}\label{eq:subProblem_minvolWl_splitted}
    \begin{aligned}
    \text{find} \quad \arg\min_{W,Z}
& \quad   G(W, \tilde W) + \lambda  \revise{D_{KL}}(Z, \bar{W}) + \alpha
 g(W,\tilde W) + \mathcal{I}_{\mcalX_W}(W)\\
                                      &\text{s.t. } W - Z = 0,
    \end{aligned}
\end{equation}
where $\mathcal{I}_{\mcalX_W}(W)$ denotes the indicator function associated to the convex set $\mcalX_W=:\{W \in \mathbb{R}^{m \times r_{l-1}} | W^\top e = e, W \geq 0 \}$, $\alpha=\frac{\alpha_\ell}{\lambda_\ell}$, and $\lambda=\frac{\lambda_{\ell+1}}{\lambda_\ell}$ .
The augmented Lagrangian function associated to Problem \eqref{eq:subProblem_minvolWl_splitted} is
\begin{equation}\label{eq:AugLagrangian_subprobinW}
    \begin{aligned}
        L_\rho(W,Z,U) :=  & G(W, \tilde W) + \lambda \revise{D_{KL}}(Z, \Bar{W}) + 
\alpha
 g(W,\tilde W)  + \mathcal{I}_{\mcalX_W}(W) + \frac{\rho}{2} \|W - Z +  U \|_F^2 , 
    \end{aligned}
\end{equation} 
where $U$ denotes the \textit{scaled} dual variables associated to constraints $W-Z=0$, written in the so-called scaled form.
Here-under, we detail the iterative procedure to compute the solution $(W,Z)$ for Problem \eqref{eq:subProblem_minvolWl_splitted}. As for classical ADMM methods, each iteration of our procedure performs three steps. Given the current iterates $(W^i, Z^i, U^i)$, the three steps are:
\begin{enumerate}
    \item $W$-minimization:
    \begin{equation}\label{eq:Step1ADMM}
       \begin{aligned}
           W^{i+1} & := \underset{W \in \mcalX_W}{\argmin} \Big\{   G(W, \tilde W) +  \alpha 
 g(W,\tilde W) + \frac{\rho}{2} \|W - Z^i + U^i \|_F^2 \Big\},
       \end{aligned}
    \end{equation}
    where $\tilde W$ is the current iterate of the main algorithm. 
    \item $Z$-minimization:
    \begin{equation}\label{eq:Step2ADMM}
        \begin{aligned}
            Z^{i+1} & := \underset{Z}{\argmin} \{ \lambda \revise{D_{KL}}(Z, \Bar{W}) + \frac{\rho}{2}\|W^{i+1} + U^i - Z \|_F^2\}.
        \end{aligned}
    \end{equation}
    \item Dual Updates: 
    \begin{equation}
        U^{i+1}:=U^i + W^{i+1} - Z^{i+1}.
    \end{equation}
    
\end{enumerate}
It remains to present how we tackle Problems \eqref{eq:Step1ADMM} and \eqref{eq:Step2ADMM}. 
\paragraph{W-minimization}
The update of $W$ is computed based on the methodology recently introduced in \cite{Leplat2021}. 
We obtain the following updates in matrix form: 
\begin{equation}\label{eq:minvol_WlUpdate}
    \hat{W}(\mu) = \tilde{W} \odot 
    \frac{\Big[\big[[C+e \mu^\top]^{.2}+S\big]^{.\frac{1}{2}}-(C + e \mu^\top)\Big]}{[T]}, 
\end{equation}
where $A = (\tilde{W}^{\top}\tilde{W} + \delta I)^{-1}$,  $A^+=\max(A,0) \geq 0$, $A^-=\max(-A,0) \geq 0$,  $C=e_{m,r_\ell-1}H^\top - 4 \alpha (\tilde{W} A^-) - \rho (Z^i - U^i)$, $T=4 \alpha \tilde{W}(A^+ + A^-) + 2\rho e_{m,r_\ell}$, $S=(8 \alpha \tilde{W} (A^+ + A^-) + 4 \rho e_{m,r_\ell} )\odot \left( \frac{[Y]}{[\tilde{W}H]}H^\top \right)$, and $e_{m,r_\ell-1}$ and $e_{m,r_\ell}$ are respectively $m$-by-$r_\ell-1$ and $m$-by-$r_{\ell}$ matrices of all ones, and $\mu \in \mathbb{R}^{r_\ell}$ denotes the vector of Lagrange multipliers associated to equality constraints on the columns of $W$, see \cite{Leplat2021} for further details. 
One can easily observe that updates defined in Equation \eqref{eq:minvol_WlUpdate} satisfy the nonnegativity constraints, given $\tilde{W} \geq 0$. Moreover, as per \cite[Proposition~2]{Leplat2021}, the constraints $\hat{W}(\mu)^\top e = e$ are satisfied for a unique $\mu^\star \in \mathbb{R}^{r_\ell}$. The computation of  $\mu^\star$ is achieved by using a Newton-Raphson procedure for solving $\hat{W}(\mu)^\top e = e$. The update $W^{i+1}$ for Problem \eqref{eq:Step1ADMM} is finally performed using Equation \eqref{eq:minvol_WlUpdate} with $\mu=\mu^\star$.

\paragraph{Z-minimization}
For this step, one can observe that the objective function to minimize is separable w.r.t. each entry of factor $Z$, hence each entry can be optimized independently. The updates are obtained by computing $\hat{Z}$ which satisfy the first-order optimality conditions, a.k.a the KKT conditions. For $\beta=1$, and denoting by $z$ the $(o,p)$-th  entry of $Z$ of interest, we are looking for $\hat{z}$ 
such that 
$$
\nabla_{z} \{ \lambda \revises{D}_{KL}(z,\Bar{W}_{o,p}) + \frac{\rho}{2} \|z - V^{i}_{o,p} \|_2^2 \}_{z=\hat{z}}= 0,  
$$ 
\revises{where $V^{i} = W^{i+1} + U^i$.}  
It boils to solve the following scalar equation in $z$:
\begin{equation}\label{eq:scalar_Zmin}
    \log(z) + b + \nu z = 0, 
\end{equation}
where $b=-\log(\Bar{W}_{o,p})-\nu V^{i}_{o,p} $ and $\nu=\frac{\rho}{\lambda}$.
Equation \eqref{eq:scalar_Zmin} can be solved in closed-form:
\begin{equation}\label{eq:sol_scalarEq_Zmin}
    \hat{z} = \frac{\mathcal{W}(e^{-b} \nu)}{\nu}, 
\end{equation}
where $\mathcal{W}(.)$ denotes the Lambert $\mathcal{W}$-function. Equation \eqref{eq:sol_scalarEq_Zmin} can be easily expressed in matrix form; the update $Z^{i+1}$ for Problem \eqref{eq:Step2ADMM} finally writes:
\begin{equation}
    Z^{i+1} = \frac{[\mathcal{W}(e^{.-B} \nu)]}{\nu}
\end{equation}
where $B=-\log(\Bar{W}) - \nu V^{i}$ and the $\log(.)$ is applied element-wise to the matrix $\Bar{W}$. 

Finally, the three steps of the ADMM-like procedure detailed above for solving Problem \eqref{eq:subProblem_minvolWl} are repeated either for a maximum number iteration $i_{max}$ or until the stopping criterion $\|W^{i} - Z^{i} \|_F \leq \epsilon$ is reached, with $\epsilon$ a threshold defined a priori. 




Algorithm~\ref{minvolADKL} summarizes our proposed algorithm for minimum-volume deep $\beta$-NMF in the case $\beta = 1$, that is, for minimum-volume deep KL-NMF (recall that $W_0 = X$ and $W_\ell$ has $r_\ell$ columns with $r_0 = n$).

\renewcommand{\baselinestretch}{1}
\algsetup{indent=1em}
\begin{algorithm}[!htbp]
\caption{Algorithm for Minimum-Volume Deep KL-NMF \label{minvolADKL}}
\begin{algorithmic}[1] 
\REQUIRE Input data matrix $X$, number of layers $L$, inner ranks $r_l$, weight parameters $\lambda_{\ell}$ and $\alpha_{\ell}$, a scalar $\delta > 0$, a maximum of iterations $i_{max}$, a threshold $\epsilon$, and parameter $\rho > 0$ for ADMM procedure.
\ENSURE An approximate solution to Problem \ref{eq:ourmodel}
    \medskip  
\FOR{$k=1,...$}
    \FOR{$l\in\{1,...,L\}$} 
        \STATE \emph{\% Update of factors $H_\ell$}
        \STATE $H^k_{\ell} \leftarrow H^{k-1}_{\ell} \odot \frac{\left[W_\ell^{T,k-1} \left( \frac{[W^{k}_{\ell-1}]}{[W^{k-1}_{\ell}H^{k-1}_{\ell}]}\right)\right]}{[W_{\ell}^{T,k-1} e_{m \times r_{\ell-1}}]}$
        \STATE \emph{\% Update of factors $W_\ell$}
        \IF{$l<L$}
            \STATE \emph{\% ADMM-procedure}
             \STATE $\tilde W \leftarrow W_\ell^{k-1}$, $A \leftarrow (\tilde{W}^\top \tilde W + \delta I)^{-1}$, $A^+ \leftarrow \max(A,0)$, $A^- \leftarrow \max(-A,0)$, 
             \STATE $T \leftarrow 4 \frac{\alpha_\ell}{\lambda_\ell} \tilde W(A^+ + A^-) + 2\rho e_{m,r_\ell}$, $S \leftarrow (8 \frac{\alpha_\ell}{\lambda_\ell} \tilde W (A^+ + A^-) + 4 \rho e_{m,r_\ell} )\odot \left(\frac{[W^k_{\ell-1}]}{[\tilde W H_\ell^k]}(H^{k}_\ell)^\top\right)$ 
            \STATE $i \leftarrow 1$
            \STATE $W^0 \leftarrow W_\ell^{k-1}$, $Z^0 \leftarrow W^0$, $U^0 \leftarrow 0$
            \STATE $\nu \leftarrow \frac{\rho}{\lambda_{\ell+1}}$
            \WHILE{$i \leq i_{max}$ \& $\|W^i - Z^i\|_F\leq \epsilon$}
            \STATE \emph{\% W-minimization}
            \STATE $V^i \leftarrow Z^i - U^i$
            \STATE $C \leftarrow e_{m,r_\ell-1}(H^{k}_\ell)^\top - 4 \frac{\alpha_\ell}{\lambda_\ell} (\tilde W A^-) - \rho V^i$
            \STATE $\mu \in \text{root}(\hat{W}(\mu)^\top e =e)$ over $\mathbb{R}^{r_\ell}$
            \STATE $W^{i} \leftarrow \tilde W \odot 
    \frac{\Big[\big[[C+e \mu^\top]^{.2}+S\big]^{.\frac{1}{2}}-(C + e \mu^\top)\Big]}{[T]}$
            \STATE \emph{\% Z-minimization}
            \STATE $V^i \leftarrow W^{i+1} + U^i$, $B \leftarrow -\log(W^{k-1}_{\ell+1} H^{k-1}_{\ell+1}) - \nu V^{i}$
            \STATE $Z^{i+1} \leftarrow \frac{[\mathcal{W}(e^{.-B} \nu)]}{\nu}$
            \STATE \emph{\% Dual Updates}
            \STATE $U^{i} \leftarrow U^{i-1} + W^{i} - Z^{i}$
            \STATE $i \leftarrow i + 1$
            \ENDWHILE
            \STATE $W_\ell^k \leftarrow W^i$
        \ELSIF{$l=L$}
            \STATE \emph{\% Update from \cite{Leplat2021}}
        \ENDIF
        

    \ENDFOR
\ENDFOR
\RETURN $\{W_L,H_L,H_{L-1},...,H_1\}$
\end{algorithmic}  
\end{algorithm} 
\renewcommand{\baselinestretch}{1}

\section{Numerical Experiments} \label{sec:appli} 

In this section, we report  the use of deep \revise{$\beta$-NMF} for three applications: 
facial feature extraction, 
topic modeling, and 
hyperspectral unmixing. The codes are written in MATLAB R2021a and available from 
\begin{center}
    \url{https://github.com/vleplat/deep-beta-NMF-public}
\end{center} 
and can be used to reproduce all experiments described below. 

\revise{It is important to acknowledge the challenge of quantitatively assessing the performance of our novel deep $\beta$-NMF models. First, we are the first to introduce deep NMF models based on the $\beta$-divergences with $\beta \in \big\{0,1,\frac{3}{2}\big\}$.} 
Second, the absence of ground-truth real-world data for deep NMF models further complicates performance evaluation, as pointed out in the survey paper by \cite{de2021survey}. \revise{For these reasons, we will only use 3 and 4 layers, while providing quantitative comparisons between deep NMF and multilayer NMF only in terms of objective function values.}  
A promising avenue for future research lies in the development of datasets specifically designed for deep NMF models. These datasets would facilitate more robust and empirical evaluations, addressing this critical need in the field.

\revise{
\subsection{Facial feature extraction with $\beta$-NMF, 
$\beta = 3/2$}\label{subsec:facialExtraction}

Let us apply deep $\beta$-NMF with $\beta=\frac{3}{2}$ on the CBCL face data set used in the seminal paper of Lee and Seung who introduced NMF to the machine learning community~\cite{Lee1999Learning}. 
It contains 2429  faces, each with $19 \times 19$ pixels. 
Note that $\beta$-NMF with $\beta = 3/2$ has been shown to perform well for imaging tasks; see~\cite{fevotte2014nonlinear}.  It also allows us to show that the updates developed in Appendix~\ref{sec:append} for the case $\beta=\frac{3}{2}$ perform well. 
We chose the penalty parameters $\lambda_\ell$ as in~\cite{de2023consistent}, that is, the $\lambda_\ell$ are chosen such that each term in the objective is equal to one another at initialization, that is, 
all values $\lambda_\ell \revise{D_{3/2}}(W_{\ell-1}^{(0)}, W_\ell^{(0)} H_\ell^{(0)})$ are equal to one, where $\big(W_1^{(0)},W_2^{(0)},\dots,H_\ell^{(0)}\big)$ is the initialization. This is the default in our implementation, with $\lambda_\ell = \frac{1}{\revise{D_{3/2}}(W_{\ell-1}^{(0)}, W_\ell^{(0)} H_\ell^{(0)})}$. 
The data matrix, $X \in \mathbb{R}^{2429 \times 361}$, contains vectorized images on its rows. As it is now well established, NMF, with $X \approx WH$, is able to extract facial features as the rows of $H$, such as eyes, noses and lips; see Figure~\ref{fig:Msvsdeepnmfcbclfacialfeatures}. 

Let us now apply multilayer and $\beta$-NMF with $\beta=\frac{3}{2}$ on this data set with four layers, and $r = [80, 40, 20, 10]$. 
For each layer of multilayer $\beta$-NMF, we run 1000 iterations of the standard MU. 
We initialize deep $\beta$-NMF with 500 iterations of multilayer $\beta$-NMF, and then run it for 500 iterations using our proposed Algorithm~\ref{ADKL}; see Appendix~\ref{sec:append} for the updates. This means that deep $\beta$-NMF is initialized with the solution of multilayer $\beta$-NMF obtained after only 500 iterations. 
We repeat this experiment 35 times, and 
Figure~\ref{fig:Msvsdeepnmfcbcl} reports the median of the evolution of the error of the four layers of deep $\beta$-NMF, \revise{that is, $D_{3/2}(W_{\ell-1},W_\ell H_\ell)$},  divided by the final error obtained by multilayer $\beta$-NMF after 1000 iterations. As expected, these ratios are initially 
larger than~1, since deep $\beta$-NMF is initialized with multilayer $\beta$-NMF after only 500 iterations. 
\begin{figure}[ht!]
\begin{center}
\includegraphics[width=9cm]{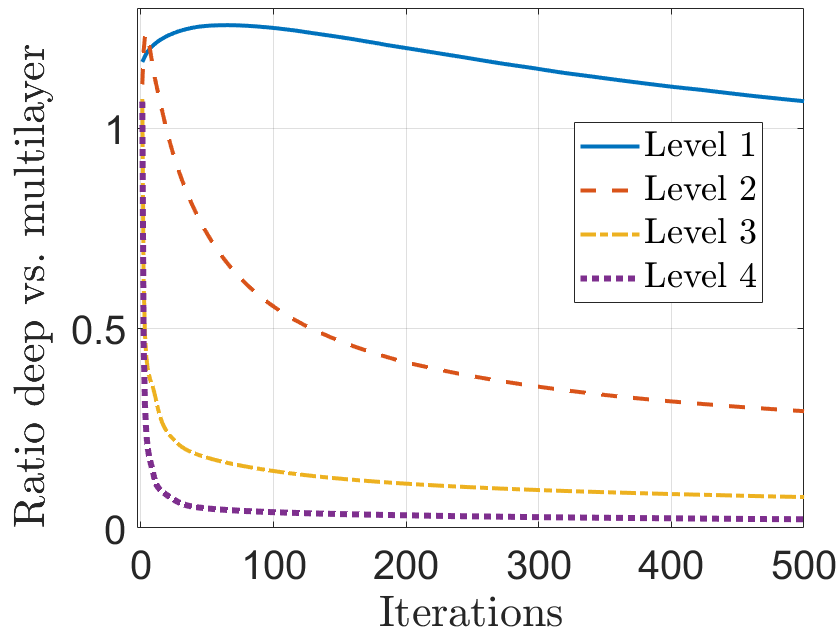} 
\caption{Evolution of the median errors at the different levels of deep $\beta$-NMF with $\beta=\frac{3}{2}$ (initialized with multilayer $\beta$-NMF after 500 iterations) divided by the error of multilayer $\beta$-NMF after 1000 iterations.} 
\label{fig:Msvsdeepnmfcbcl}
\end{center}
\end{figure}

The main observation from Figure~\ref{fig:Msvsdeepnmfcbcl} is the following: Because deep $\beta$-NMF needs to balance the error between each layer, the error at the first layer remains larger than that of multilayer $\beta$-NMF, which is expected. However, the errors at the next three layers becomes quickly  significantly smaller; see also the first column of Table~\ref{tab:rescbcl}. 
At convergence, the error at the second (resp.\ third and fourth) layer is on average about 3.5 (resp.\ 13 and 47) times smaller than that of  multilayer $\beta$-NMF.  

\begin{center}
\begin{table}[ht!]
\begin{center} 
\begin{tabular}{|c|c|c|c|}
\hline 
   & error deep/ML (in \%) 
   & sparsity ML (in \%) 
   & sparsity deep (in \%)   \\  \hline
Layer 1 & 106.7 $\pm$   0.9  &   57.3  $\pm$  0.5 &  69.6  $\pm$  0.5 \\  
Layer 2 & 29.2  $\pm$   0.9  &    33.9 $\pm$  0.7            & 53.6  $\pm$  0.8   \\  
Layer 3 & 7.7  $\pm$   0.4  &         18.8  $\pm$  0.8          & 37.4 $\pm$   1.2 \\  
Layer 4 & 2.1   $\pm$   0.3  &         10.2  $\pm$  0.8          & 17.6 $\pm$   1.1 \\
\hline
\end{tabular} 
\end{center}
\caption{Deep vs.\ Multilayer (ML) $\beta$-NMF with $\beta=\frac{3}{2}$: average and standard deviation for the error of deep $\beta$-NMF divided by that of ML $\beta$-NMF (\revise{first} column), and average and standard deviation for the sparsity of the facial features of ML  $\beta$-NMF (second column) and  deep $\beta$-NMF (third column). }
\label{tab:rescbcl} 
\end{table}
\end{center}

Table~\ref{tab:rescbcl} also reports the average sparsity of the facial features at each layer, using the widely used Hoyer sparsity~\cite{hoyer2004non} given by 
\[
\text{sparsity}(x) \; = \; \frac{\sqrt{n} - \frac{||x||_1}{||x||_2}}{\sqrt{n} - 1} \; \in \; [0,1], 
\]  
for an $n$-dimensional vector $x$. This measure is equal to one \revise{if} $x$ has a single non-zero entry, and is equal to zero if all entries of $x$ are equal to one another. 

First, observe that for both deep and multilayer $\beta$-NMF, the sparsity decreases as the factorization unfolds: this is unavoidable since facial features at deeper levels are nonnegative linear combinations of facial features at shallower levels. For example, the facial features at the second layer are given by $H_2 H_1$, that is, nonnegative linear combinations of the facial features at the first level, $H_1$; see also the discussion in Section~\ref{sec:ident_deepNMF}. 
Second, we observe that deep $\beta$-NMF produces significantly sparser facial features (namely +12.3\% at layer 1,  +19.7\% at layer 2, +18.6\% at layer 3, +7.4\% at layer~4). This makes sense because deep $\beta$-NMF balances the error at the four layers, and sparser features at the first layer gives more degree of freedom to generate features at the next layers. In fact, a dense feature at the first layer can only generate denser ones at the next layers. This is an interesting side result of deep $\beta$-NMF: it can be used to solve sparse NMF, without parameter tuning.  

\begin{figure}[ht!]
\begin{center}
\begin{tabular}{cc}
Multilayer - layer 1 & Deep - layer 1 \\ 
  \includegraphics[width=0.49\textwidth]{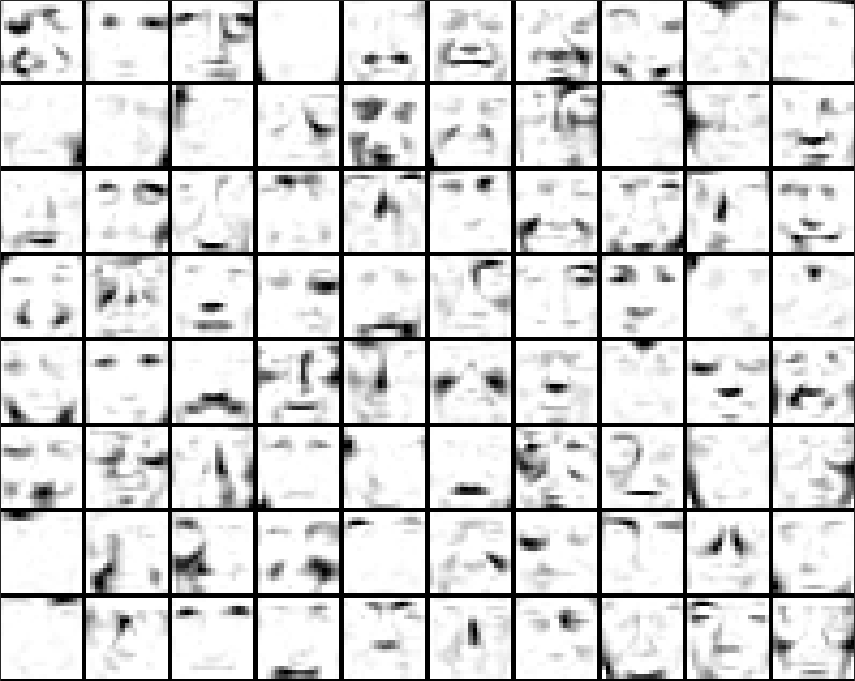} 
    &    
\includegraphics[width=0.49\textwidth]{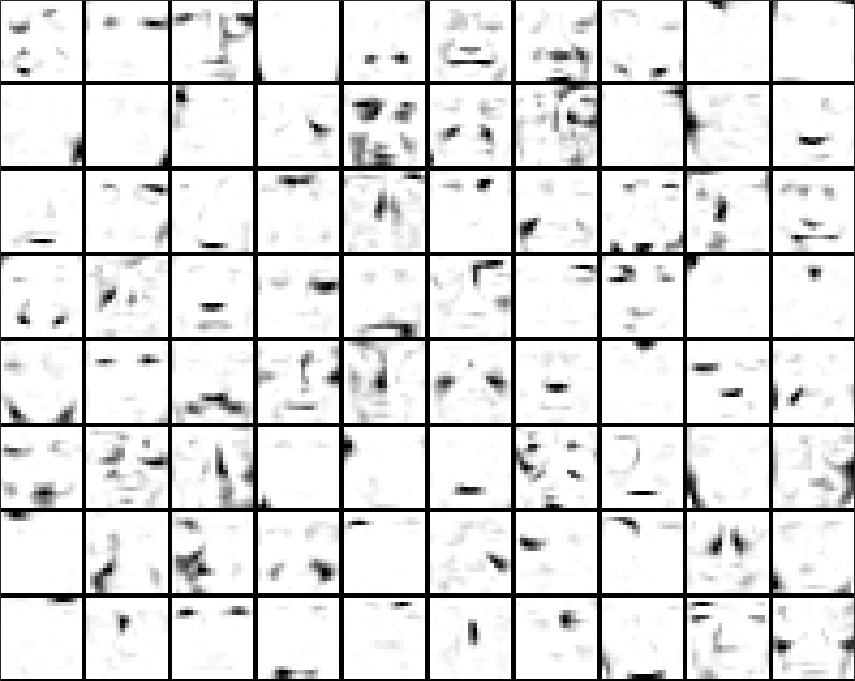} \\ 
Multilayer - layer 2 & Deep - layer 2 \\
  \includegraphics[width=0.49\textwidth]{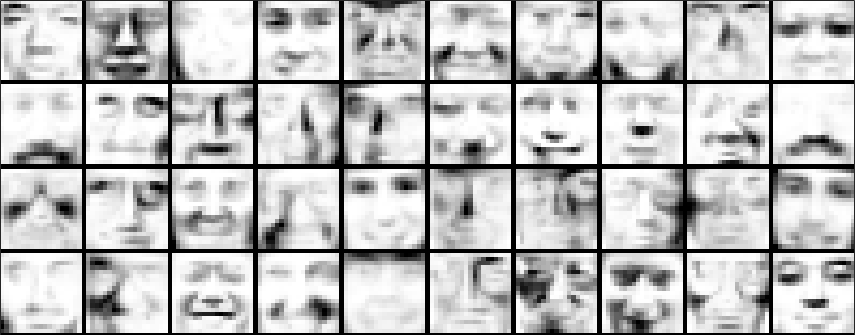} 
    &    
\includegraphics[width=0.49\textwidth]{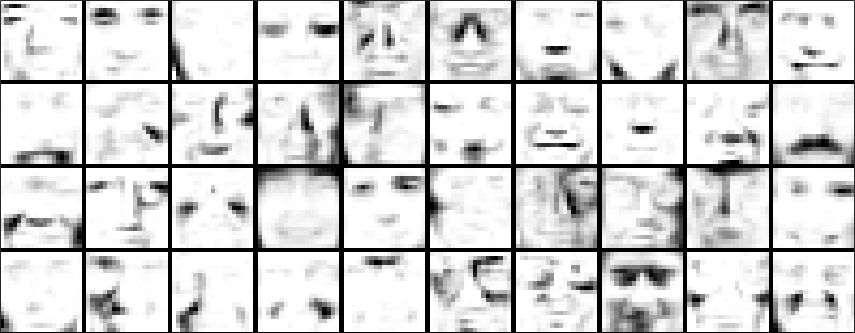} \\ 
Multilayer - layer 3 & Deep - layer 3 \\ 
  \includegraphics[width=0.49\textwidth]{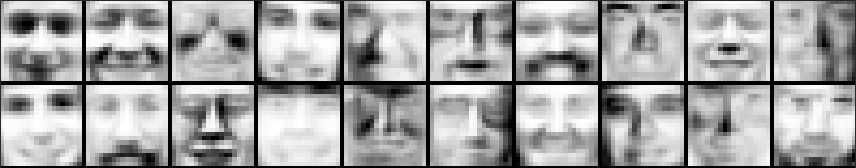} 
    &    
\includegraphics[width=0.49\textwidth]{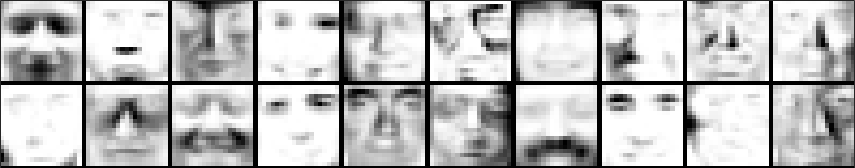} \\
Multilayer - layer 4 & Deep - layer 4 \\ 
  \includegraphics[width=0.49\textwidth]{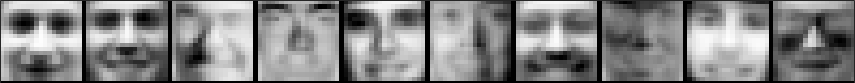} 
    &    
\includegraphics[width=0.49\textwidth]{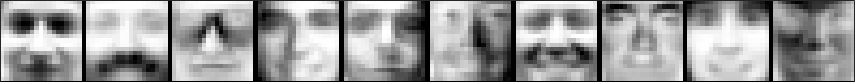}
\end{tabular}
\caption{Example of facial features extracted by multilayer $\beta$-NMF vs.\  deep $\beta$-NMF for $\beta=\frac{3}{2}$. } 
\label{fig:Msvsdeepnmfcbclfacialfeatures}
\end{center}
\end{figure} 
Figure~\ref{fig:Msvsdeepnmfcbclfacialfeatures} displays the facial features of multilayer and deep NMF at the different layers (for the last run of our experiment, since it does not make sense to average facial features). 
We observe that most of the facial features of the first layer of deep and multilayer $\beta$-NMF are similar, the main  difference is that those of deep $\beta$-NMF are sparser. However, starting at the second at layer 2, 3 and 4, where the error of deep $\beta$-NMF is significantly smaller, some facial features are completely different (e.g., the second one), showing that deep $\beta$-NMF produces  different outcomes than multilayer $\beta$-NMF. 

We provide additional results for this data set in Appendix~\ref{sec:append_2} with $\beta=1$ and with three layers, which lead to similar conclusions as the ones detailed above.

}






\subsection{Topic modeling} 

Topic modeling aims to discover the underlying topics or themes in a collection of documents. It is a form of unsupervised learning that can help organize, summarize, and understand large textual datasets.
In topic models, one typically assumes that a document is generated by a mixture of topics, each of which is a distribution over words in the vocabulary; see, e.g.,~\cite{churchill2022evolution} and the references therein.   
Topic modeling models aim to represent each document through the learned topics and understand the topics in the corpus through the
most probable words of each topic. 

NMF has been successfully used in this context, initiated by the paper of Lee and Seung~\cite{Lee1999Learning}. 
If $X$ is a word-by-document matrix, its NMF, $X \approx WH$, extract topics in $W$ which is a word-by-topic matrix, while $H$ allows one to classify the documents across the topics.  
In this context, deep NMF allows one to extract layers of topics: from lower level topics to higher level ones (e.g., tennis and football belong to sports). We will see an example below; see also~\cite{will2023neural} for a different but similar deep NMF model used for topic modeling. 
It has been well established that the KL divergence is more appropriate for the analysis of document data set. The reason is that such data sets are sparse (most documents only use a few words in the dictionary). In fact, the KL divergence amounts for a Poisson counting process; see, e.g.,~\cite[Chapter 5]{gillis2020nonnegative} and the references therein. 

In this section, we apply deep KL-NMF to the TDT2-top30 dataset and compare its performance with multilayer NMF. 
The TDT2 corpus (Nist Topic Detection and Tracking
corpus) consists of data collected during the first half of 1998 and taken
from 6 sources, including 2 newswires (APW, NYT), 2 radio programs
(VOA, PRI) and 2 television programs (CNN, ABC). Only the largest 30
categories are kept, thus leaving us with 9394 documents in total~\cite{Cai_2008}. 
We ran the experiment with three layers with $r=[20, 10, 5]$. Moreover, since deep NMF is computationally more intensive, we preprocess the data set to keep only the most important words. To do that, we perform a rank-20 NMF of $X$, and keep the 30 most important words in each topic (that is, each column of $W$); we used the code from \url{https://gitlab.com/ngillis/nmfbook/}. 

We use a similar setting as for the CBCL dataset, except that we only run the algorithms once (we will focus on the analysis of the topics obtained) and use $\lambda = [4, 2, 1]$, that is, we give more importance to the first layers. Otherwise, we observe the first layer topics  were getting too similar: for example, giving too much importance to the term $\|W_1 - W_2 H_2\|_2$ will make $W_1$ become rank deficient (since $r_2 < r_1$), and hence its columns will become more colinear. 

\begin{figure}[ht!]
\begin{center}
\includegraphics[width=10cm]{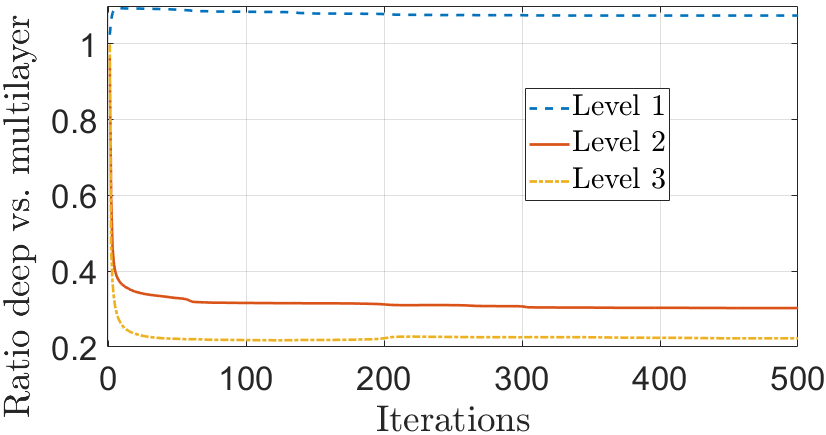} 
\caption{Evolution of the error at the different levels of deep KL-NMF  divided by the error of multilayer KL-NMF.} 
\label{fig:Msvsdeepnmftdt2}
\end{center}
\end{figure}

Figure~\ref{fig:Msvsdeepnmftdt2} reports the evolution of the error of deep KL-NMF divided by the final error obtained by multilayer KL-NMF, exactly as for the CBCL data set. 
As for CBCL, the errors of deep KL-NMF at the first level is slightly larger (namely 107.5\%), while the error at the second and third levels are significantly smaller (namely 30\% and 22\%, respectively) than that of multilayer NMF. 

Let us try to analyse the topics extracted by deep KL-NMF, and the relationships between them. 
For each column of $W_\ell$, which contain the topics, we sort the words in order of importance, and report them in that order. 
Table~\ref{tab:topicslayer1} provides the 10 most important words for the second layer, and Table~\ref{tab:topicslayer3_deep}  provides the 5 most important words for the third layer. For simplicity, we do not report the topics at layer 1 (there are 20), but a similar analysis can be made, and they are available from our code online. 

\begin{center}
\begin{sidewaystable}[!htbp]
\begin{center} \setlength{\tabcolsep}{0.3pt} 
\begin{tabular}{|c|c|c|c|c|c|c|c|c|c|}
\hline 
 Topic 1: & Topic 2: & Topic 3: & Topic 4: & Topic 5: & Topic 6: & Topic 7: & Topic 8: & Topic 9: & Topic 10: \\
  {\bf  Asian}  &  {\bf Tobacco}  & {\bf American }  & {\bf Nuclear  } 
    &  {\bf Stock } & {\bf Media } & {\bf Asian}  & {\bf Olympic}   & {\bf Clinton-Lewinsky } & {\bf Iraq}  \\
     {\bf politics} &  {\bf bill} &  {\bf politics}  &  {\bf conflicts}  &   {\bf market} &   &  {\bf economy} & {\bf games } & {\bf scandal} &  {\bf  war} \\ 
  \hline
united&   tobacco&   president&   nuclear&   percent&   spkr&  economic&  olympic& lewinsky&   iraq\\
states&    industry&   clinton&  india &   market &  says&  crisis &  city&  president& weapons\\
president&   defense&   house &   pakistan &   stock &   reporter&  billion&   games&  
 spkr &  united\\
indonesia&   bill&   political &   government&   economic &   news&   financial&   team & clinton&  iraqi\\
suharto&   companies&   government&   party&   bank&   game&   economy &   olympics& white&   states\\
general&  smoking&   china &  president&    asia &   correspondent&   asian& nagano&  house& saddam\\
american&    military&    administration &  tests&  prices &  going &  government &  gold& starr&  miliatary\\
government&  legislation&   white&   minister&   asian&   headline &   asia&  won& lawyers& inspectors\\
indonesian&  health&  visit&  indias&   economy &   voa & bank&  women & jury& baghdad\\
jakarta& officials& rights& political& israel & reports& south &  police & case &security
\\
\hline 
\end{tabular} 
\end{center}
\caption{Most important words in each topic of the second layer of deep KL-NMF.}
\label{tab:topicslayer1} 
\end{sidewaystable}
\end{center}

\begin{center}
\begin{table}[!htbp]
\begin{center} \setlength\tabcolsep{4.5pt}
\begin{tabular}{|c|c|c|c|c|}
\hline 
 Topic 1 & Topic 2& Topic 3& Topic 4 & Topic 5\\
 {\bf Politics} &  {\bf Military conflicts} & {\bf Economy} & {\bf Clinton-Lewinsky} &  {\bf Mixed topics}\\
  \hline
president& iraq &   percent & spkr & tobacco\\
clinton& united & economic & lewinsky & olympic\\
government& states & crisis & president & team\\
nuclear& weapons& economy & clinton & city\\
india & iraqi & asian & white & games\\
house & military& financial & house & going\\
political& saddam & billion & starr & olympics\\
china& security& market & lawyers &won\\
pakistan& inspectors& bank & news & nagano\\
party&  president& asia& case& gold \\
\hline
 Merged topics & Merged topics 
 & Merged topics  & Merged topics  & 
 Merged topics \\
3 \&  4 & 1, 6, \&  10  & 5 \&  7  & 6 \& 9 & 2, 6 \& 8   \\
from layer 2 &  from layer 2 & from  layer 2 & from  layer 2 & from  layer 2 \\
 \hline
\end{tabular} 
\end{center}
\caption{Most important words in each topic of third layer of deep KL-NMF.}
\label{tab:topicslayer3_deep} 
\end{table}
\end{center}

What is interesting with deep NMF models, is that we can link the topics together, as in a hierarchical decomposition. For example, the topics of layers 2 and 3, within the columns of $W_2$ and $W_3$ respectively, are linked via the relation 
\[
W_2(:,j) \approx \sum_{k=1}^{r_3} W_3(:,k) H_3(k,j) \; \text{ for } \; j=1,2,\dots,r_2,  
\] 
where $H_3(k,j)$ tells us the importance of the $k$th topic at level 3 to reconstruct the $j$th topic at level 2. 
The last line of Table~\ref{tab:topicslayer3_deep} provides this information. Because most topics extracted at the second layer are rather different and use different words, $H_3$ is rather sparse (this in turn is because $X$ is sparse, and hence the $W_\ell$ and $H_\ell$'s are as well). 
For example, the topic 1 of the third layer (about politics) merges the topics 3 and 4 from the second layer
(about American politics and nuclear conflicts), and the 
the topic 3 of the third layer (about economics) merges the topics 5 and 7 from the second layer
(about the stock market and the Asian economic crisis). 
Except for the fifth topic of the third layer, which is a mixture of heterogeneous ones (mostly Olympic games, but mixed with the tobacco bills and the media topics), 
the other ones are rather meaningful. Interestingly, the topic at level 2 about the media (with words such as reporter, correspondent, headline, etc.) is merged into three topics where the media is present (military conflicts, political scandals, Olympic games).

\subsection{Hyperspectral imaging}


In this section, we consider hyperspectral images to evaluate the effectiveness of the proposed min-vol deep KL-NMF solved via Algorithms~\ref{minvolADKL}, in comparison to the multilayer KL-NMF~\cite{cichocki2006multilayer, cichocki2007multilayer}, and the Frobenius-norm based deep MF framework with min-vol penalty and data-centric loss function recently proposed in~\cite{de2023consistent}. 
To ease the notation, the latter will be dubbed as ``LC-DMF". All the algorithms are implemented and tested on a laptop computer with Intel Core i7-11800H@2.30GHz CPU, and 16GB memory.

\subsubsection{Data sets}\label{subsec:dataset_desImage}

\begin{figure}
     \centering
     \begin{subfigure}[b]{0.32\textwidth}
         \centering
         \includegraphics[width=\textwidth]{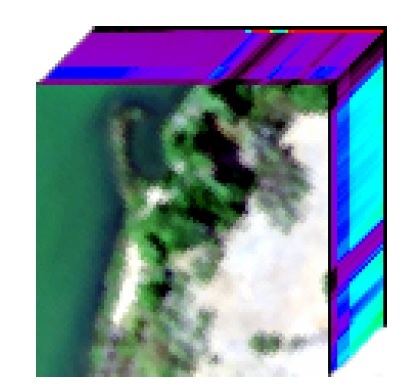}
         \caption{Samson data cube}
         \label{fig:Samsondataset}
     \end{subfigure}
     \hfill
     \begin{subfigure}[b]{0.56\textwidth}
         \centering
         \includegraphics[width=\textwidth]{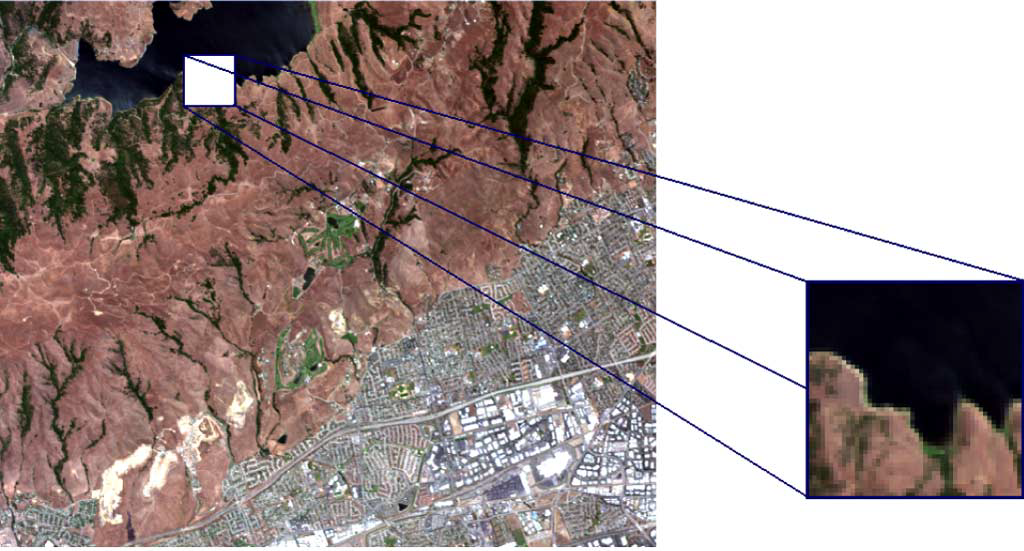}
         \caption{Moffett in synthetic colors}
         \label{fig:Moffett}
     \end{subfigure}
     \vskip -0.1in
        \caption{HSI datasets: Samson (left) -  Moffett Field acquired by AVIRIS in 1997 and the region of interest (right) represented in synthetic colors, figure reproduced from \cite{5256272}. }
        \label{fig:twoDatasets}
\end{figure}

A hyperspectral image (HSI) is an image that contains information over a wide spectrum of light instead of just assigning primary colors (red, green, and blue) to each pixel as in RGB images. The spectral range of typical airborne sensors is 380-12700 nm and 400-1400 nm for satellite sensors. For instance, the AVIRIS airborne hyperspectral imaging sensor records spectral data over 224 continuous channels. The advantage of HSI is that they provide more information on what is imaged, some of it blind to the human eye as many wavelengths belong to the invisible light spectrum. This additional information allows one to identify and characterize the constitutive materials present in a scenery. We consider the following real HSI: \begin{itemize}
    \item \textbf{AVIRIS Moffett Field}: this data set has been acquired with over Moffett Field (CA, USA) in 1997 by the JPL spectro-imager AVIRIS \footnote{\url{https://aviris.jpl.nasa.gov/data/image_cube.html} } and consists of 512$\times$614 pixels and 224 spectral reflectance bands in the wavelength range 400nm to 2500nm. Due to the water vapor and atmospheric effects, we remove the noisy spectral bands. After this process, there remains 159 bands.
    As in \cite{5256272}, we extract a 50$\times$50 sub-image from this data set, see Figure \ref{fig:Moffett}. It is widely acknowledged within the hyperspectral remote sensing community that this subimage consists of three distinct materials: vegetation, soil, and water. It is worth noting that the norm of the spectral signature for water is significantly smaller compared to the other two materials. For more detailed information on this dataset, please refer to \cite{5256272}.
    \item \textbf{Tumor}: 519 spectral bands with 13 $\times $11 pixels, corresponding to a simulated Magnetic resonance spectroscopic imaging (MRSI) of a glioma patient brain \cite{li2012simulation}. This MRSI grid contains spectra from normal tissue, as well as tumor tissue and necrosis, see \cite{li2012simulation} for more details about this data set. 
    \item \textbf{Samson}: The Samson dataset\footnote{\url{http://lesun.weebly.com/hyperspectral-data-set.html}} comprises 156 spectral bands and has a resolution of 95 $\times$ 95 pixels. It primarily contains three to four materials, namely ``Soil", ``Tree", and ``Water". In Figure \ref{fig:Samsondataset}, the hyperspectral cube is depicted, and upon closer examination, it becomes apparent that the "Soil" material actually consists of at least two sub-components, specifically sand and rocks. 
\end{itemize}


\subsubsection{Results}
In this section, we present the results obtained from the benchmarked methods for each hyperspectral imaging (HSI) dataset, as described in Section \ref{subsec:dataset_desImage}. Specifically, we showcase the abundance maps obtained for each layer of the deep models, aiming to provide a qualitative assessment of the unmixing and clustering outcomes. For all the models under consideration, we enforce a two-layer decomposition. Our objective is to achieve an accurate estimation and localization of the constitutive materials, also known as endmembers, in the first layer. The subsequent layer is expected to provide a clustering effect by merging the endmembers into more general clusters, such as vegetation vs.\ non-vegetation as illustrated in Figure \ref{fig:deepnmfurban}, or mineral vs.\ organic or healthy vs non-healthy human tissues.

To ensure completeness and reproducibility of the results, we will now provide additional details regarding the parameters that were considered for each analysis.
\begin{itemize}
    \item \textbf{AVIRIS Moffett Field}: for the AVIRIS Moffett Field dataset, we consider factorization ranks of $r_1 = 4$ and $r_2 = 2$, with a maximum number of iterations set to 300. In the deep MF framework with a min-vol penalty \cite{de2023consistent}, we impose a sum-to-one constraint on the columns of factors $W_l$. The values for penalty weights of min-vol regularizations have been tuned and set to $[2;1]$ to obtain the best results. For our proposed Algorithms~\ref{minvolADKL}, we set $\rho$ to 100, the threshold $\epsilon$ to $10^{-6}$ for the ADMM procedure and the penalty weights of minimum-volume regularizations to $[4;1]$. 
    
    \item \textbf{Tumor}: for this data set, we consider factorization ranks of $r_1 = 3$ and $r_2 = 2$, with a maximum number of iterations set to 200. As done for previous data set, we impose a sum-to-one constraint on the columns of factors $W_l$ for LC-DMF \cite{de2023consistent} with default values for min-vol penalty weights. In the case of Algorithms~\ref{minvolADKL}, the parameters for the ADMM procedure are the same as the ones considered above, whereas the penalty weights of minimum-volume regularizations are set to $[\frac{1}{4}10^{-4};10^{-2}]$ .

\item \textbf{Samson}: For this dataset, we consider factorization ranks of $r_1 = 4$ and $r_2 = 2$, with a maximum number of iterations set to 800. Again, we impose a sum-to-one constraint on the columns of factors $W_l$ for LC-DMF \cite{de2023consistent} with values for penalty weights of minimum-volume regularizations tuned and set to $[0.15;0.15]$ to obtain the best results. In the case of Algorithms~\ref{minvolADKL}, the parameters for the ADMM procedure are the same as the ones for previous data sets, whereas the penalty weights of minimum-volume regularizations are set to $[0.45;0.45]$. 
   
\end{itemize}

\begin{figure}
        \begin{tabular}{ccc}
     \includegraphics[width=0.3\textwidth]{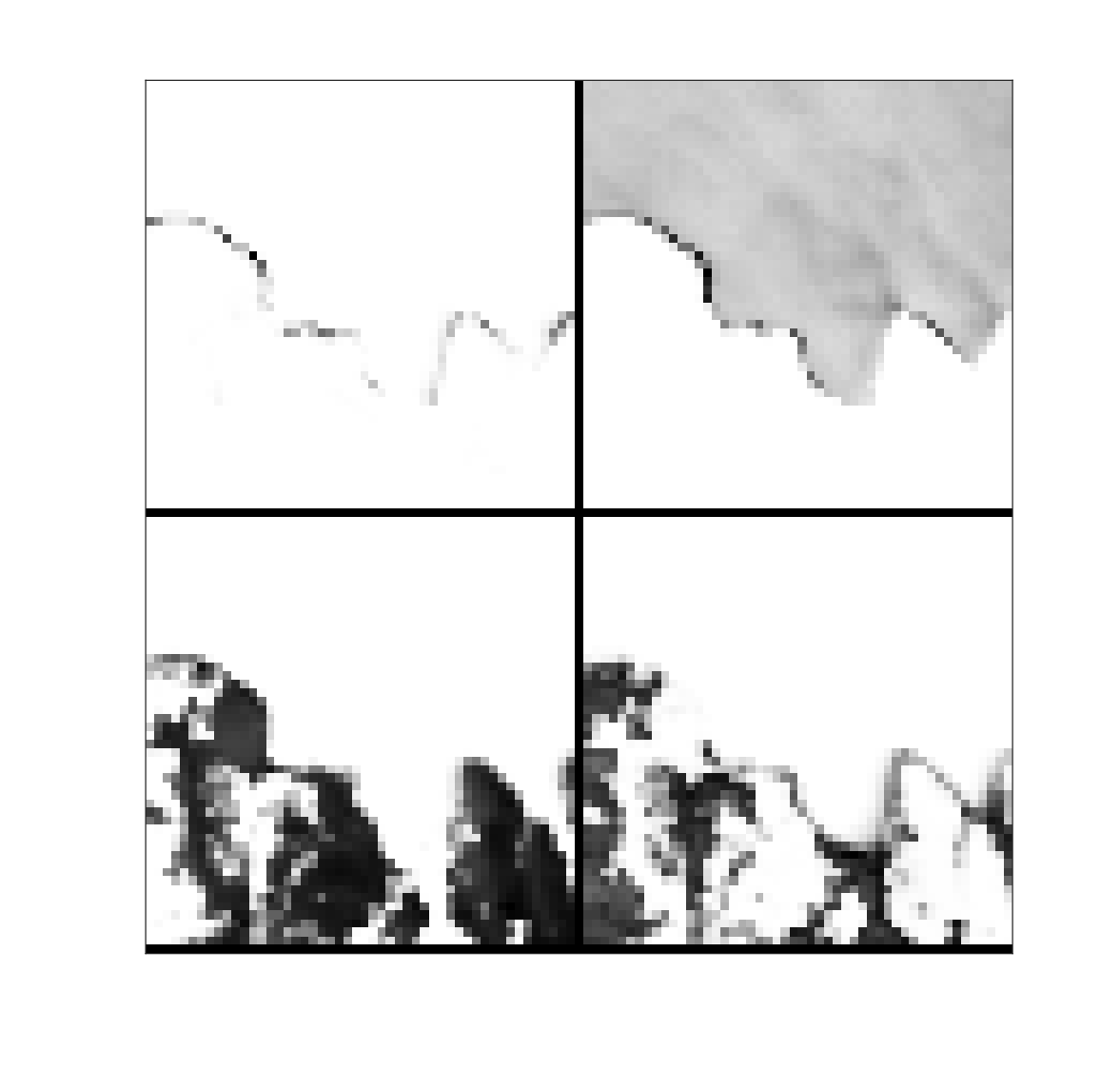}
      & 
      \includegraphics[width=0.3\textwidth]{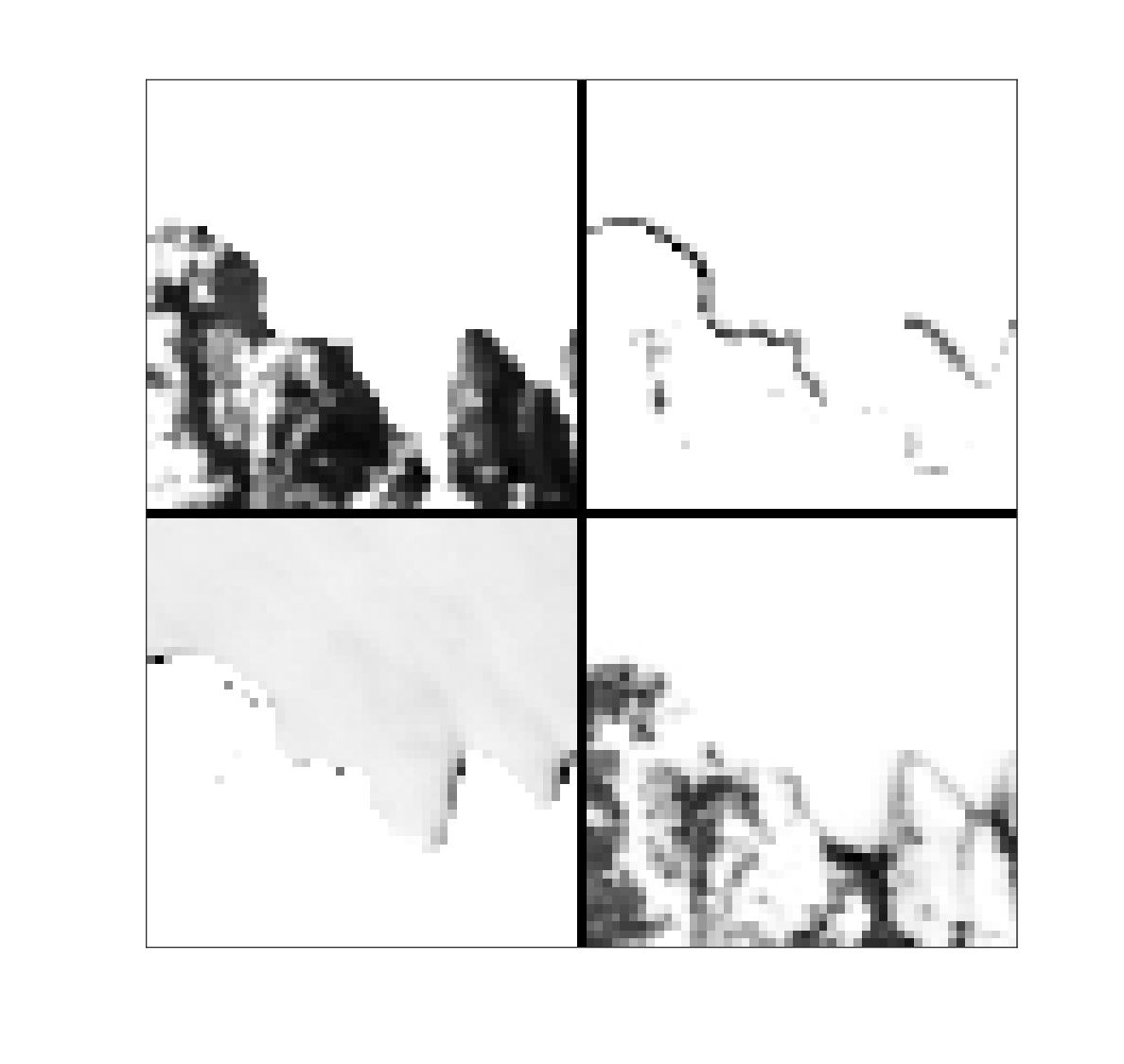}
      &  
      \includegraphics[width=0.3\textwidth]{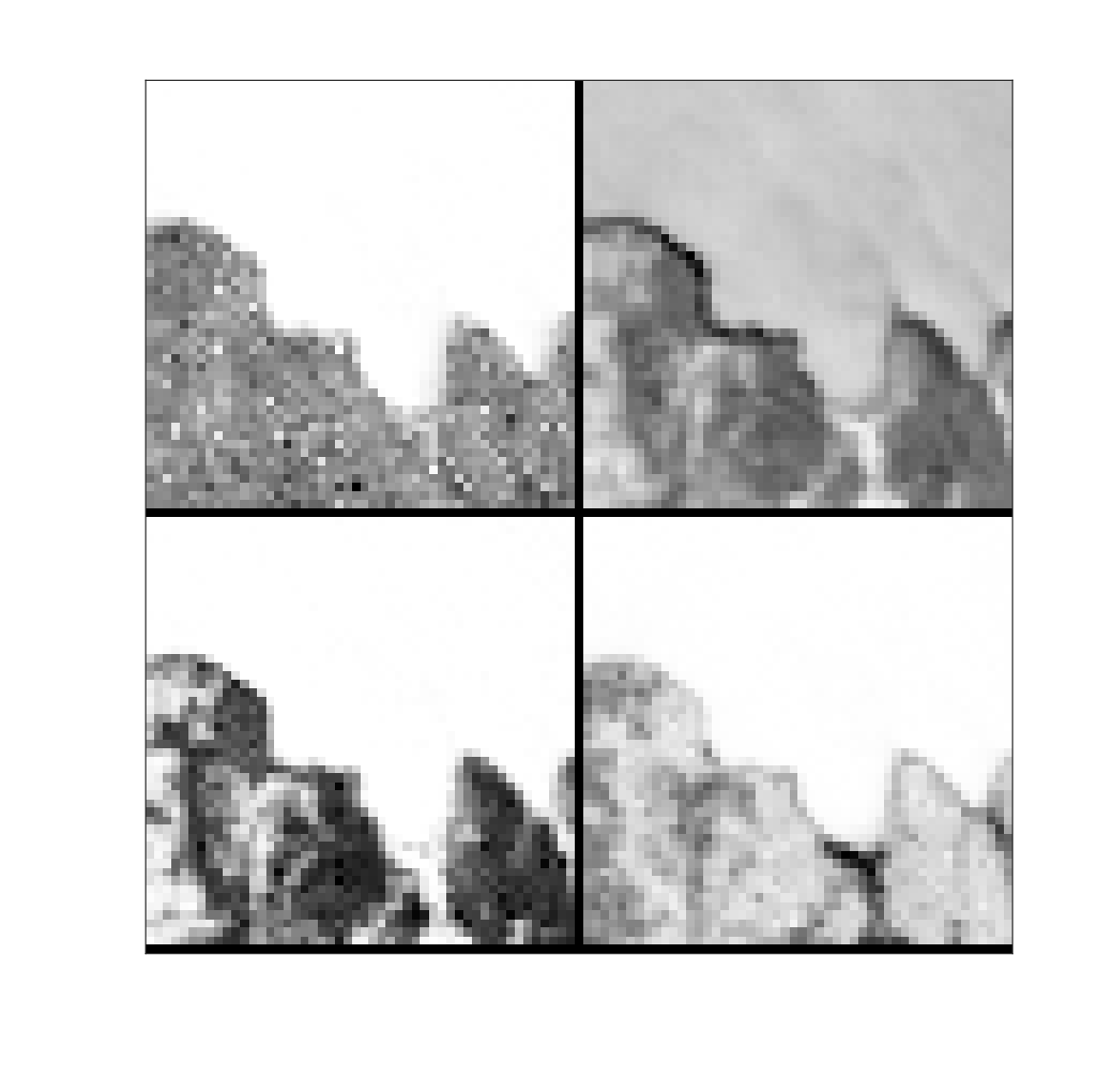} \\
         (a) deep KL-NMF 
          & (b) LC-DMF & (c) Multi-layer KL-NMF 
     \end{tabular}
        \caption{AVIRIS Moffett Field data set: From left to right abundance maps extracted from the first layer of min-vol deep KL-NMF (Alg.~\ref{minvolADKL}), LC-DMF~\cite{de2023consistent} and multi-layer KL-NMF~\cite{cichocki2006multilayer, cichocki2007multilayer}.}
        \label{fig:firstlayer_Moffett}
\end{figure}
\begin{figure}
     \begin{tabular}{ccc}
     \includegraphics[width=0.3\textwidth]{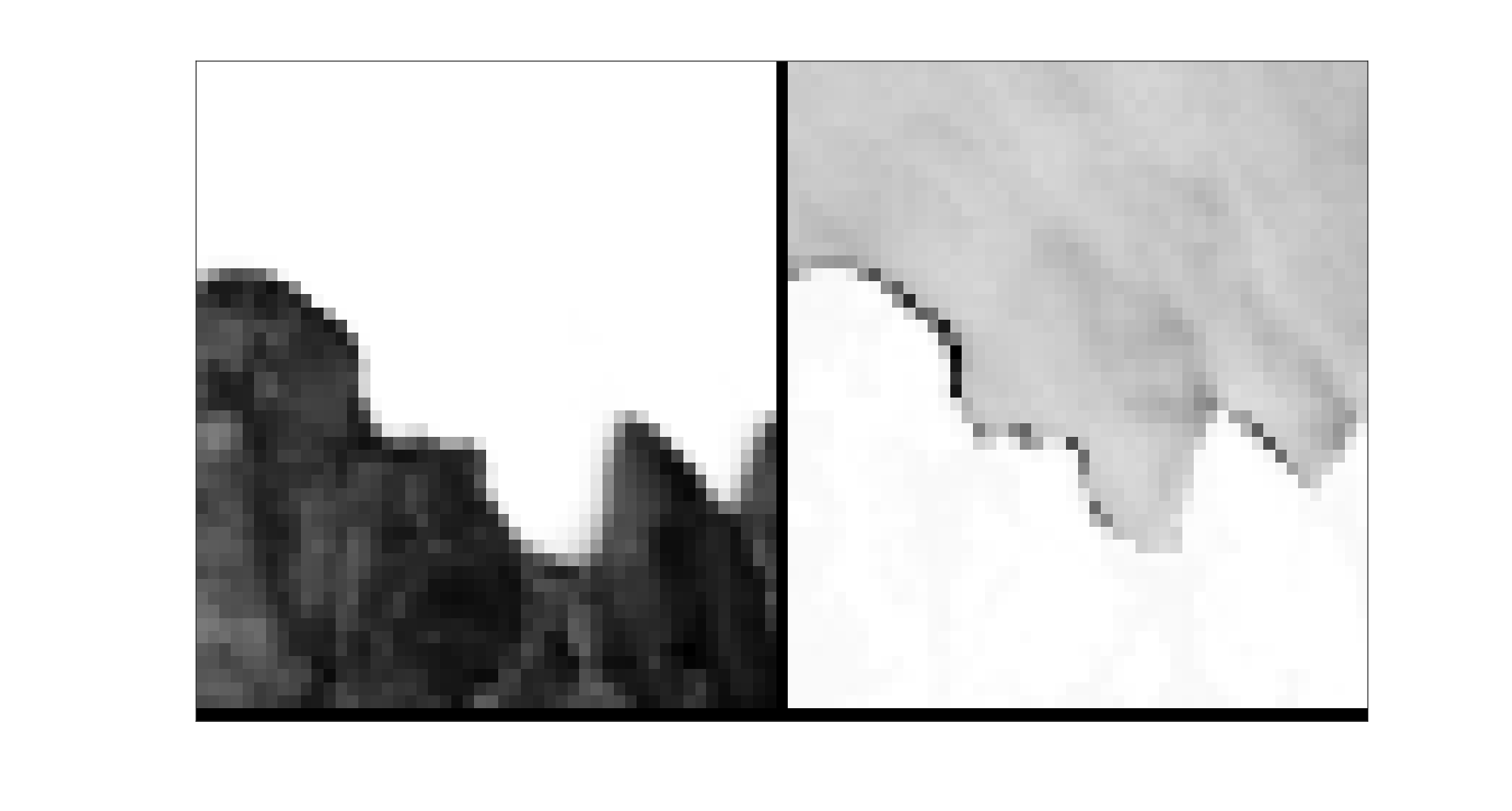} 
      & 
      \includegraphics[width=0.3\textwidth]{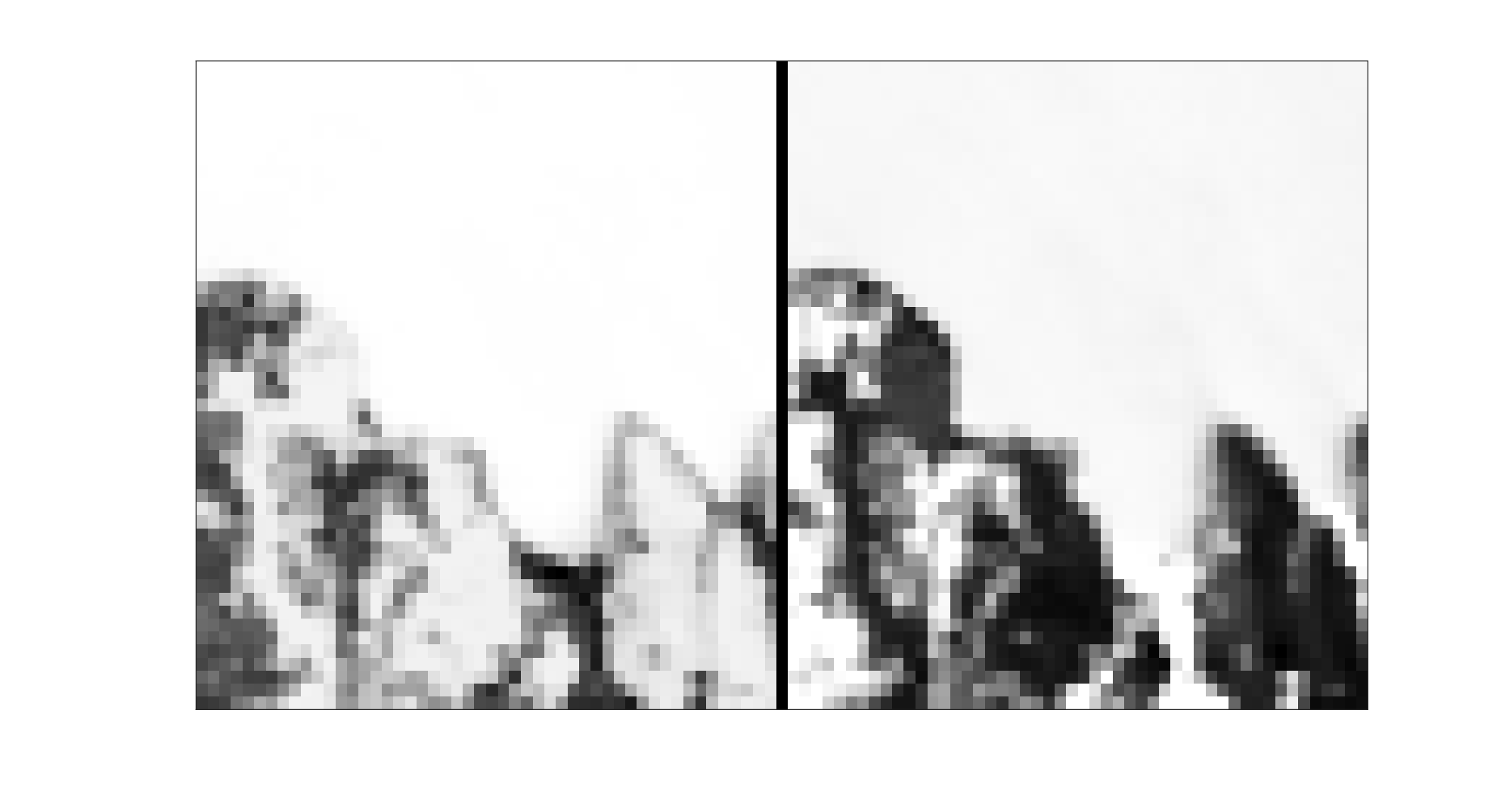}
      &  
      \includegraphics[width=0.3\textwidth]{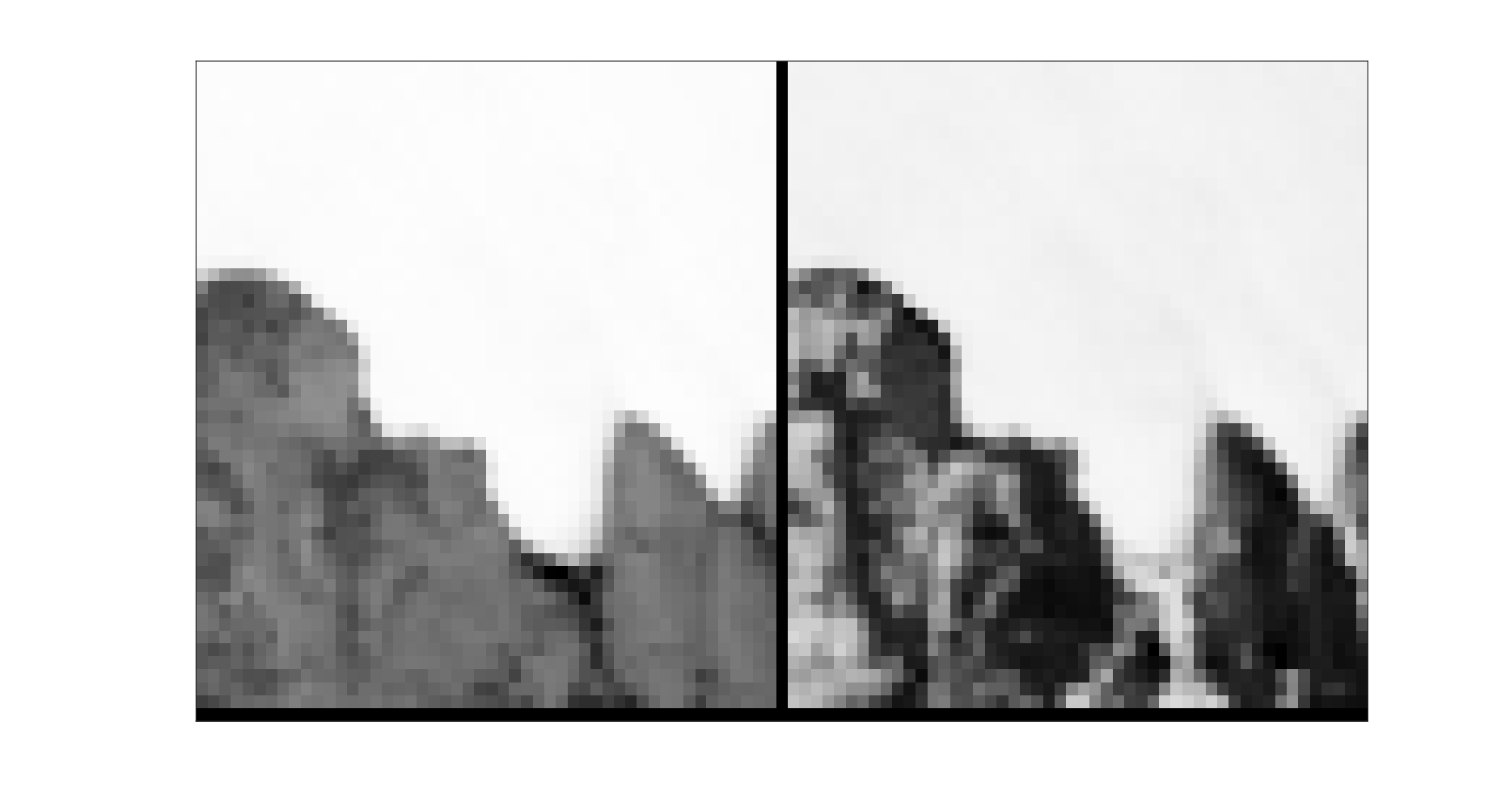} \\
         (a) deep KL-NMF 
          & (b) LC-DMF & (c) Multi-layer KL-NMF 
     \end{tabular}
        \caption{AVIRIS Moffett Field data set: From left to right abundance maps extracted from the second layer of min-vol deep KL-NMF (Alg.~\ref{minvolADKL}), LC-DMF~\cite{de2023consistent} and multi-layer KL-NMF~\cite{cichocki2006multilayer, cichocki2007multilayer}.}
        \label{fig:secondlayer_Moffett}
\end{figure}

\paragraph{Discussion on AVIRIS Moffett Field} Figure \ref{fig:firstlayer_Moffett} presents the abundance maps obtained for each layer of the three models. Impressively, both min-vol deep NMF models accurately detect the presence of water in the first layer, as well as a discernible "material" observed at the interface between the water and the soil. This interface showcases non-linear effects that arise from the phenomenon of double scattering of light. Both min-vol deep models effectively highlight these effects, with our proposed method demonstrating slightly superior accuracy in capturing such intricate features. It is worth noting that the estimation of water in this dataset is highly challenging, and the most successful results have been achieved by imposing sum-to-one constraints on the $H$ factor of NMF models, as discussed in detail in \cite{gillis2020nonnegative}. Notably, the deep KL-NMF model provides the most accurate estimation of water. Moving on to Figures \ref{fig:secondlayer_Moffett}, we observe the abundance maps obtained for the final layer of the three models. Once again, both min-vol deep KL-NMF models yield more meaningful outcomes. While LC-DMF~\cite{de2023consistent} distinguishes between vegetation and soil through clustering, min-vol deep KL-NMF (Alg.~\ref{minvolADKL}) gathers soil and vegetation, contrasting them with water.

\begin{figure}
     \centering
     \begin{tabular}{ccc}
     \includegraphics[width=0.3\textwidth]{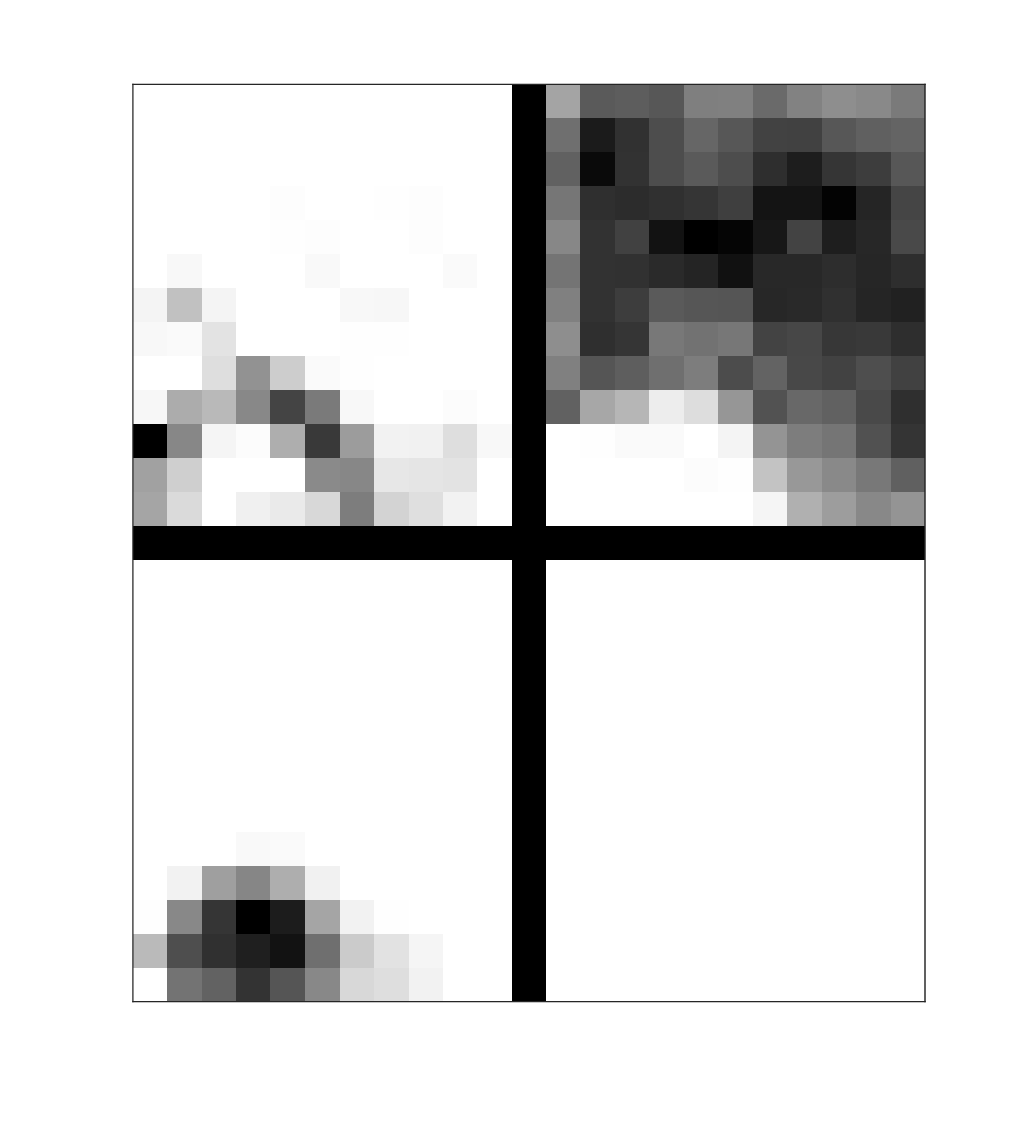}   
      & 
      \includegraphics[width=0.3\textwidth]{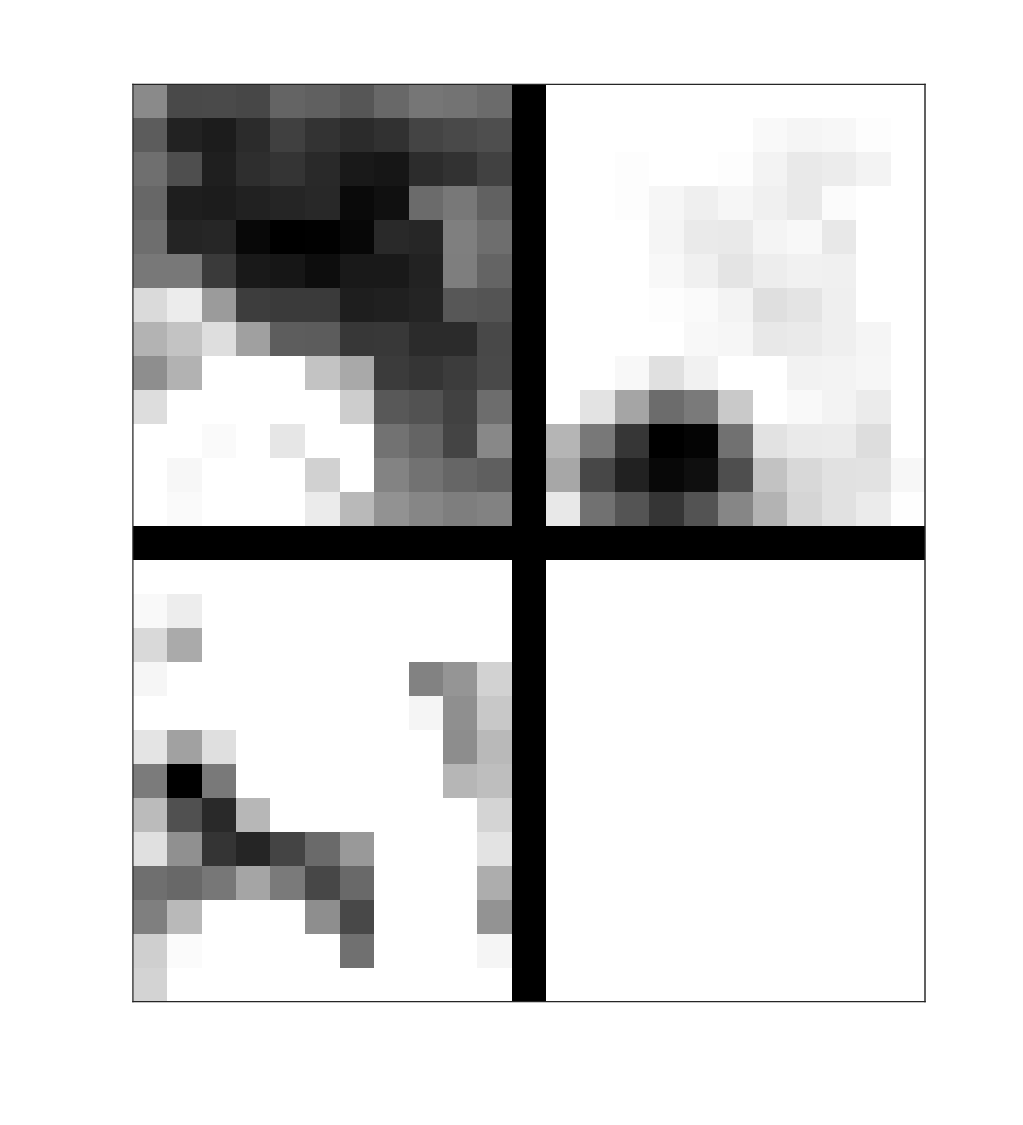}
      &  
      \includegraphics[width=0.3\textwidth]{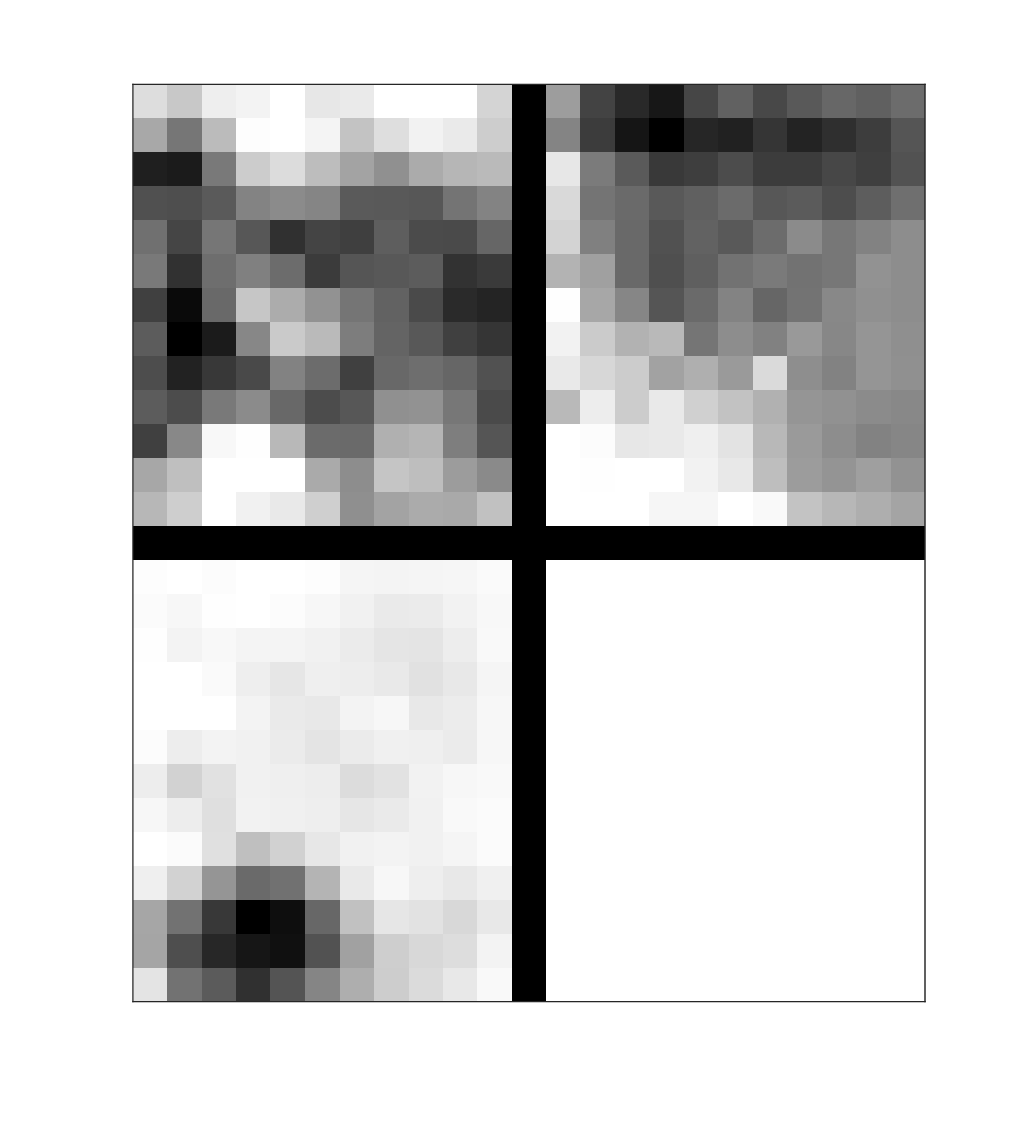} \\
         (a) deep KL-NMF 
          & (b) LC-DMF & (c) Multi-layer KL-NMF 
     \end{tabular}
        \caption{Tumor data set: From left to right abundance maps extracted from the first layer of min-vol deep KL-NMF (Alg.~\ref{minvolADKL}), LC-DMF~\cite{de2023consistent} and multi-layer KL-NMF~\cite{cichocki2006multilayer, cichocki2007multilayer}.}
        \label{fig:firstlayer_Tumor}
\end{figure}

\begin{figure}

     \begin{tabular}{ccc}
      \includegraphics[width=0.3\textwidth]{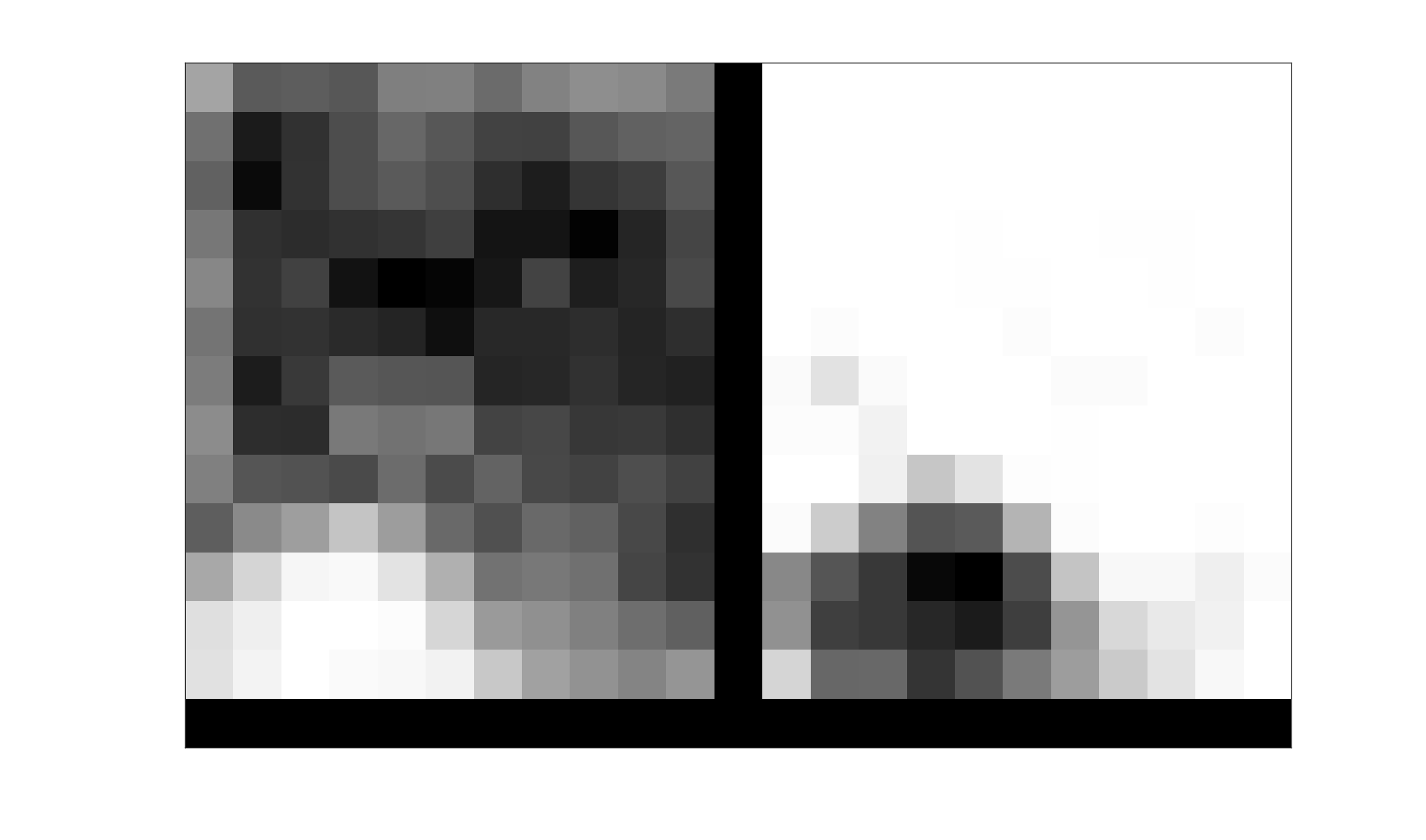}    
      & 
      \includegraphics[width=0.3\textwidth]{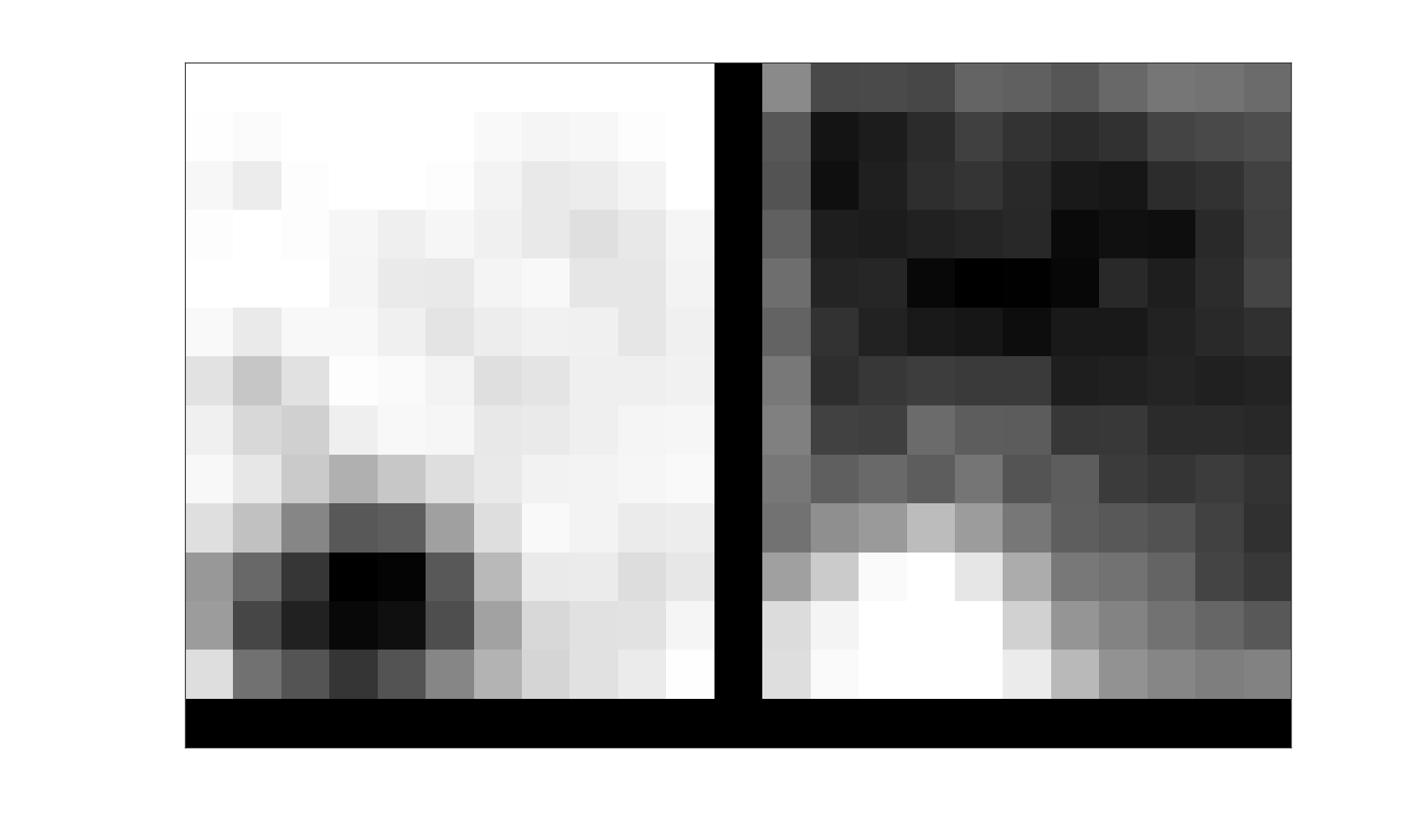}
      &  
      \includegraphics[width=0.3\textwidth]{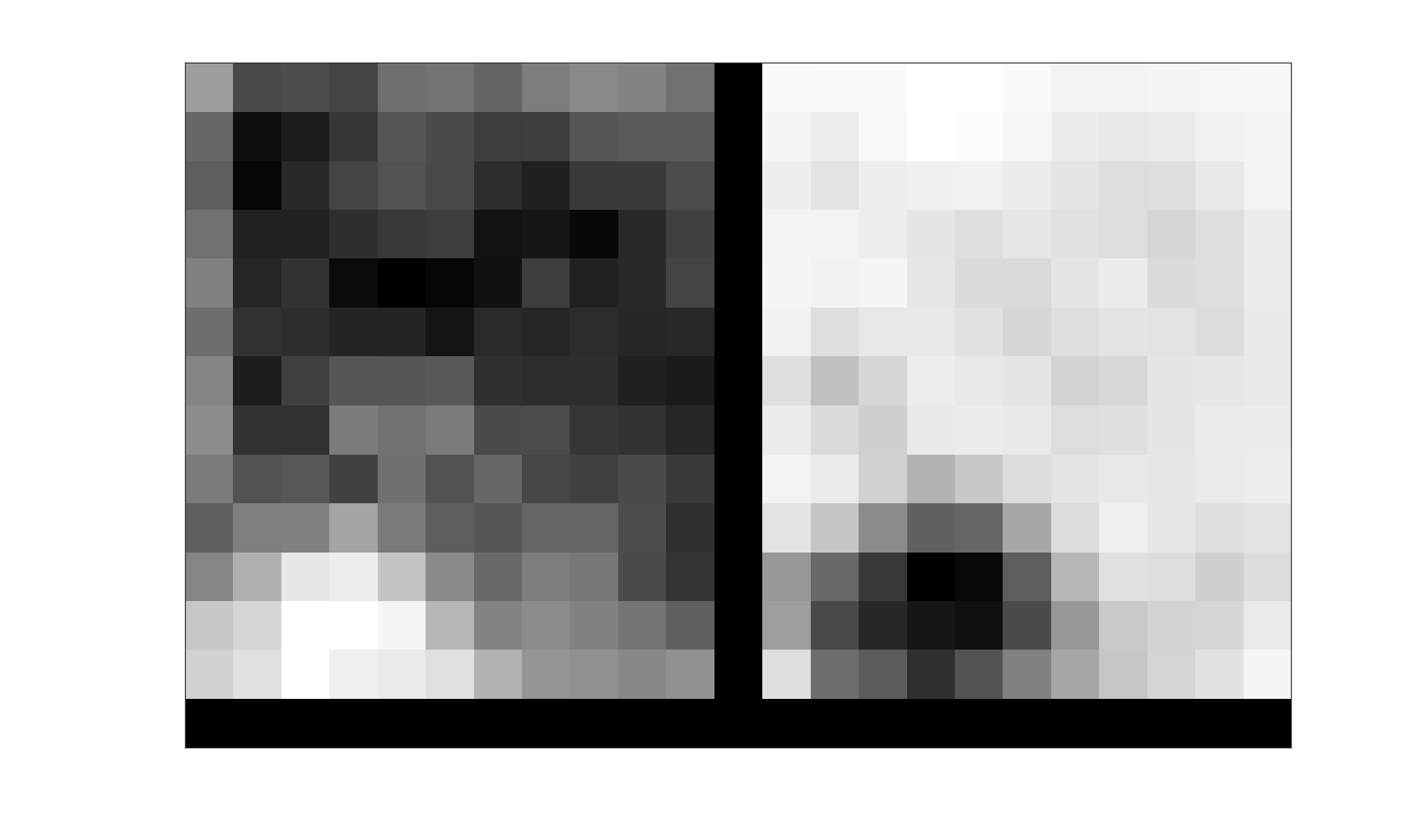} \\
         (a) deep KL-NMF 
          & (b) LC-DMF & (c) Multi-layer KL-NMF 
     \end{tabular}
        \caption{Tumor data set: From left to right abundance maps extracted from the second layer of min-vol deep KL-NMF (Alg.~\ref{minvolADKL}), LC-DMF~\cite{de2023consistent} and multi-layer KL-NMF~\cite{cichocki2006multilayer, cichocki2007multilayer}.}
        \label{fig:secondlayer_Tumor}
\end{figure}

\paragraph{Discussion on Tumor data set} Figure \ref{fig:firstlayer_Tumor} illustrates the abundance maps obtained for each layer of the three models. As reported in \cite{li2012simulation}, the dataset comprises three endmembers: the ``necrosis" forming a ball in the lower corner of the MRSI grid, the aureole-shaped tumor surrounding the necrosis, and the healthy tissue. Overall, all three models produce satisfactory results. In this analysis, it is evident that the min-vol deep NMF models more accurately extract this information, with a slightly cleaner localization achieved by min-vol deep KL-NMF (Alg.~\ref{minvolADKL}). Examining the second layers extracted by the models depicted in Figures \ref{fig:secondlayer_Tumor}, we observe that all three models exhibit a similar clustering of the endmembers extracted in the first layer, distinguishing non-healthy tissue (tumor + necrosis) from healthy tissue. Once again, min-vol deep KL-NMF showcases a slightly better separation in this regard.

\begin{figure}
     \centering
     
     \begin{tabular}{ccc}
        \includegraphics[width=0.3\textwidth]{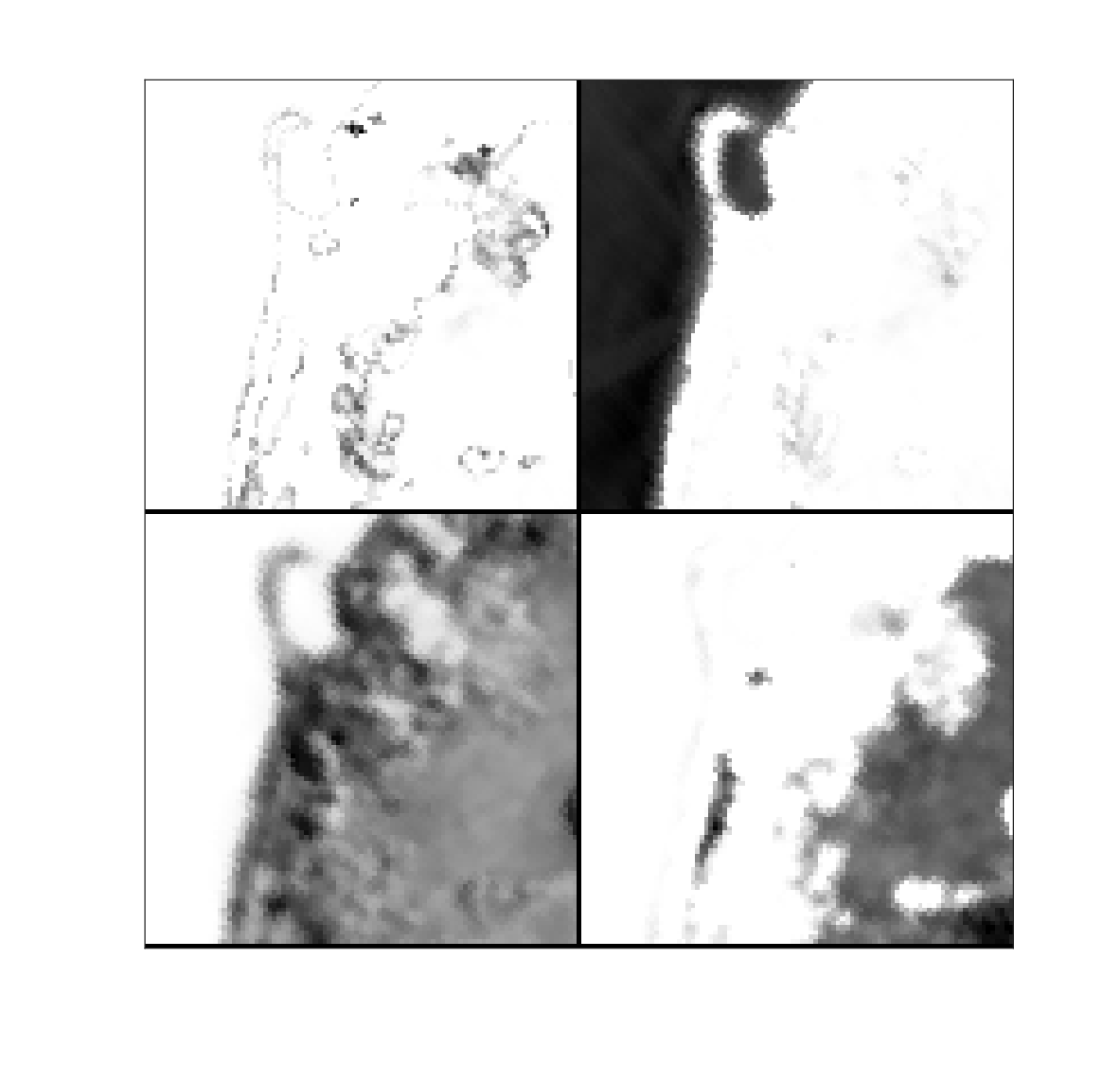}  &  
        \includegraphics[width=0.3\textwidth]{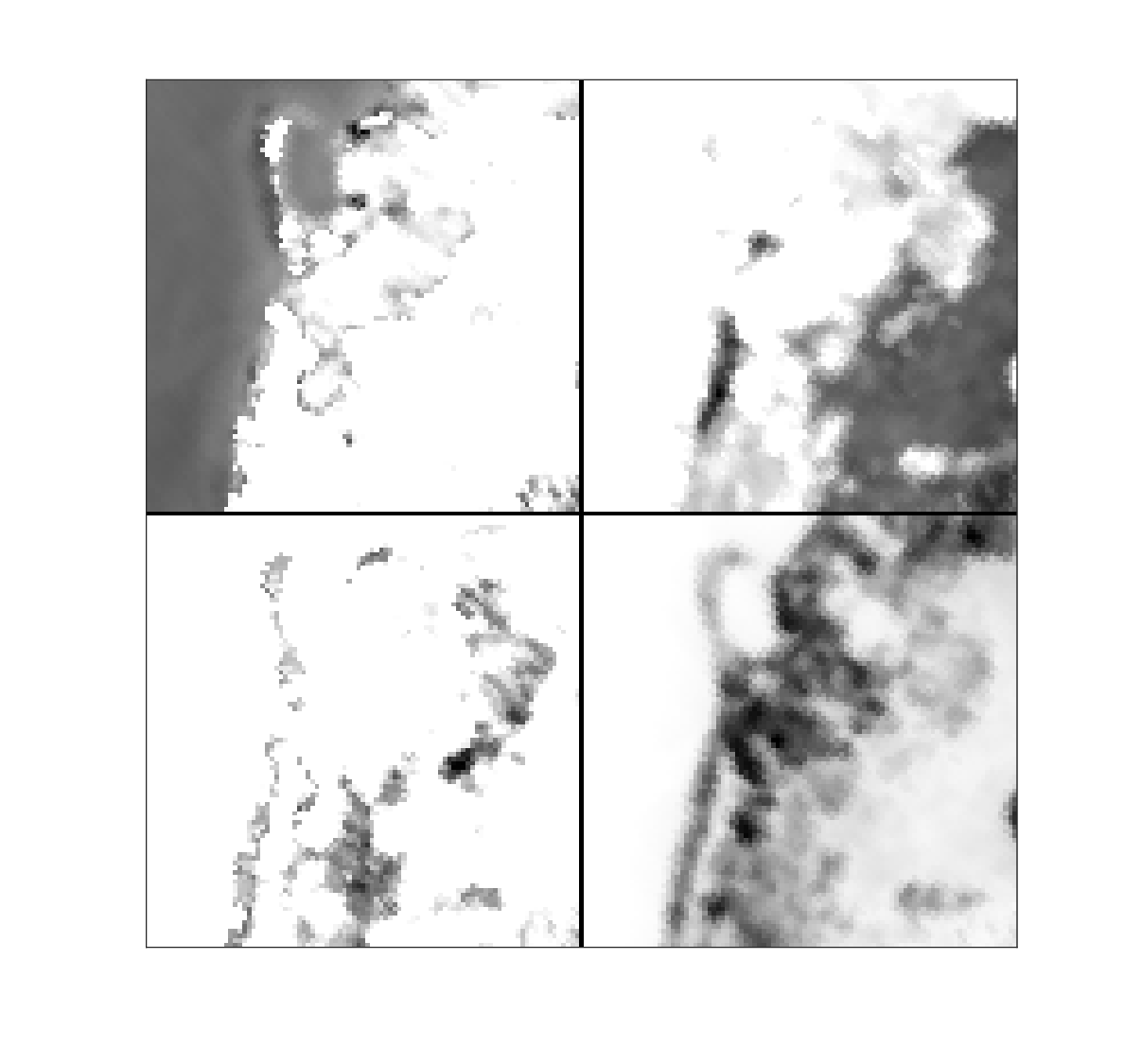}
          & 
          \includegraphics[width=0.3\textwidth]{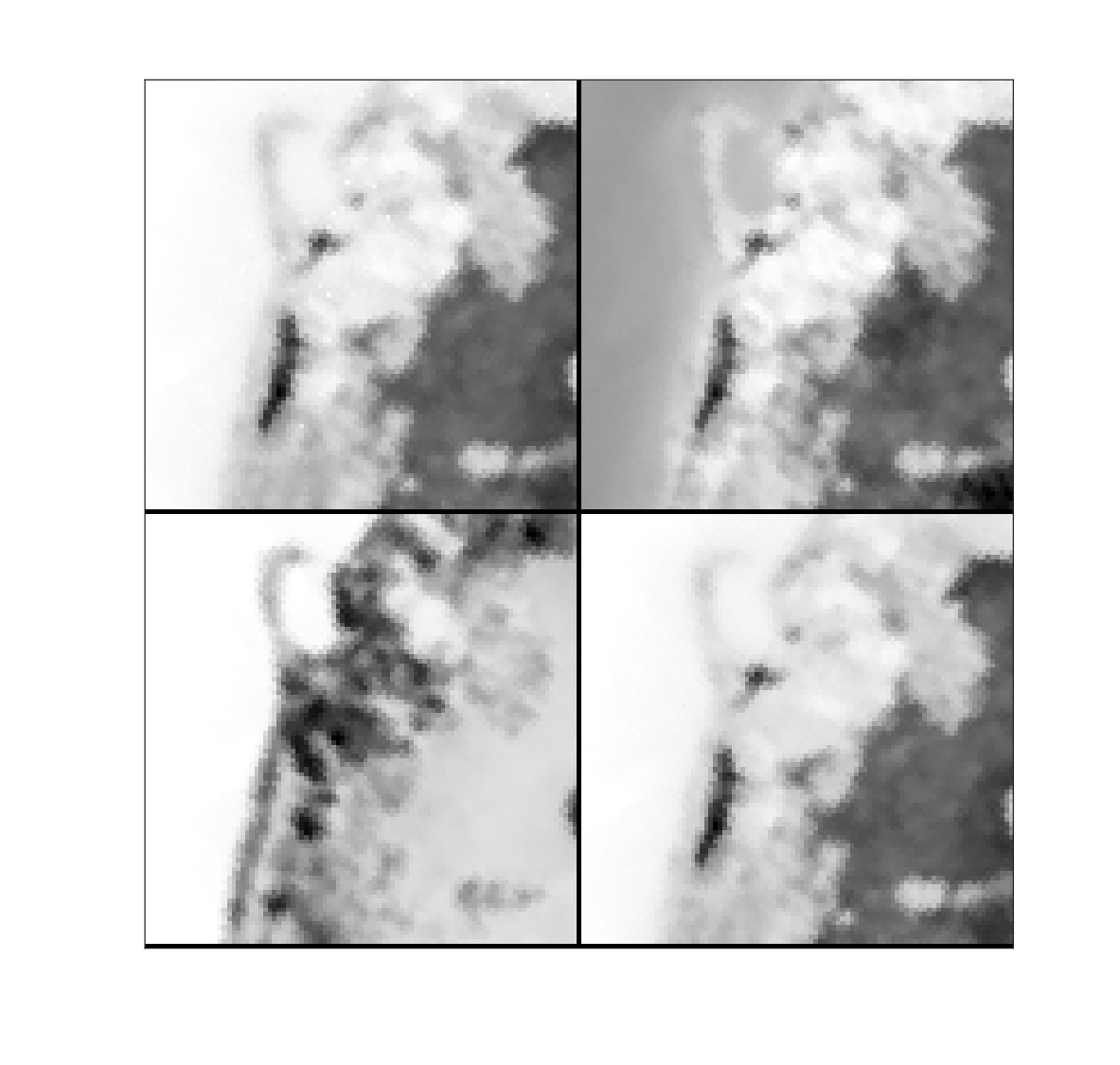} \\
          (a) deep KL-NMF (Alg.~\ref{minvolADKL})
          & 
          (b) LC-DMF~\cite{de2023consistent} 
          & 
          (c) Multi-layer KL-NMF~\cite{cichocki2006multilayer, cichocki2007multilayer} 
     \end{tabular}
        \caption{Samson data set: From left to right abundance maps extracted from the first layer of min-vol deep KL-NMF (Alg.~\ref{minvolADKL}), LC-DMF~\cite{de2023consistent} and multi-layer KL-NMF~\cite{cichocki2006multilayer, cichocki2007multilayer}.}
        \label{fig:firstlayer_Samson}
\end{figure}

\paragraph{Discussion on Samson data set}  Figure \ref{fig:firstlayer_Samson} illustrates the abundance maps obtained for each layer of the three models. Interestingly, both min-vol deep NMF models successfully extract four materials, as explained in Section \ref{subsec:dataset_desImage}: water, vegetation, soil, and additional material that likely corresponds to a second type of soil, possibly rocks. These models also capture non-linear effects at the interface between soil and water. It is worth noting that min-vol deep KL-NMF (Alg.~\ref{minvolADKL}) demonstrates better estimation of the localization of water and soil, while LC-DMF~\cite{de2023consistent} provides slightly improved separation for vegetation, resulting in fewer residuals from the soil. Moving on to Figures \ref{fig:secondlayer_Moffett}, we observe the abundance maps obtained for the final layer of the three models. Remarkably, all three deep models combine the endmembers extracted in the first layer into clusters: minerals (soil + water) versus vegetation.

\paragraph{Conclusion on hyperspectral imaging} Our proposed  min-vol deep KL-NMF (Alg.~\ref{minvolADKL}) shows promising results in hyperspectral imaging compared to two other methods. Despite its unconventional use of KL divergence instead of the Frobenius norm, deep KL-NMF exhibits superior performance in capturing intricate features, particularly in detecting water and the water-soil interface. It outperforms other methods in accurately estimating water, a challenging task in the AVIRIS Moffett Field dataset. Additionally, deep KL-NMF achieves satisfactory results in the Tumor dataset with improved localization and separation. 
In the Samson dataset, deep KL-NMF successfully extracts multiple materials and offers better localization of water and soil.


\begin{figure}
     \centering
         \begin{tabular}{ccc}
\includegraphics[width=0.3\textwidth]{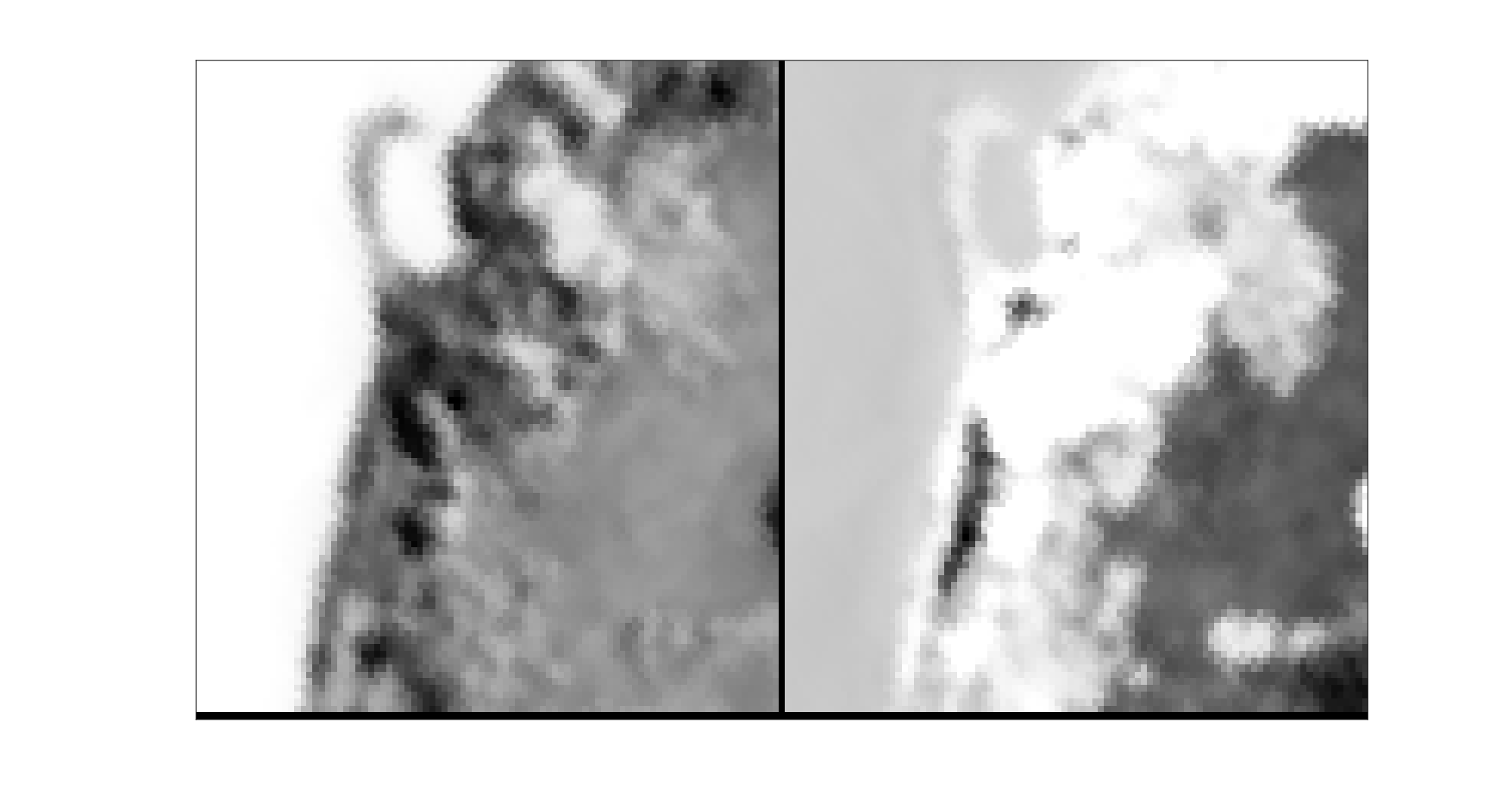}
&
\includegraphics[width=0.3\textwidth]{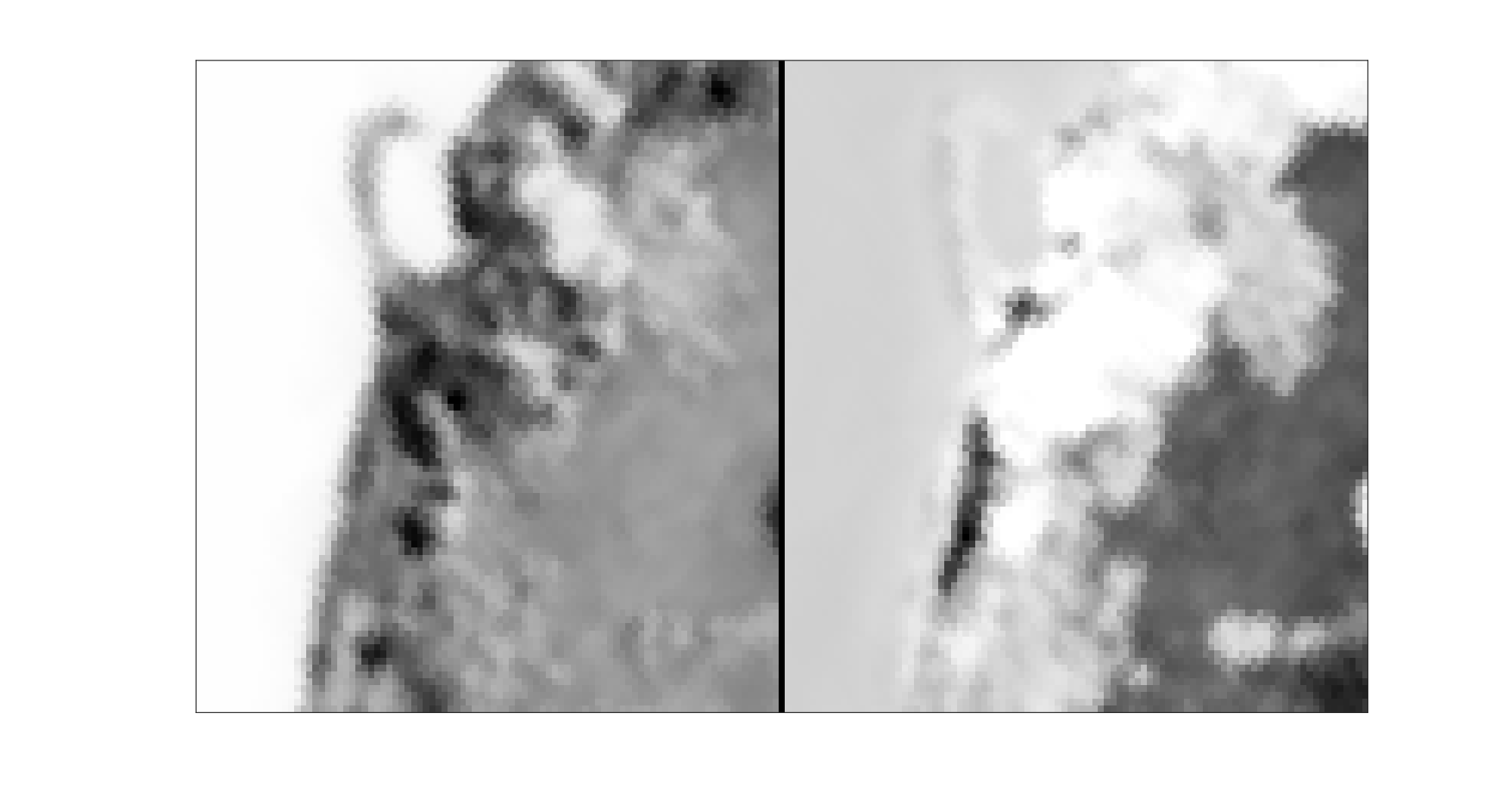}
&
\includegraphics[width=0.3\textwidth]{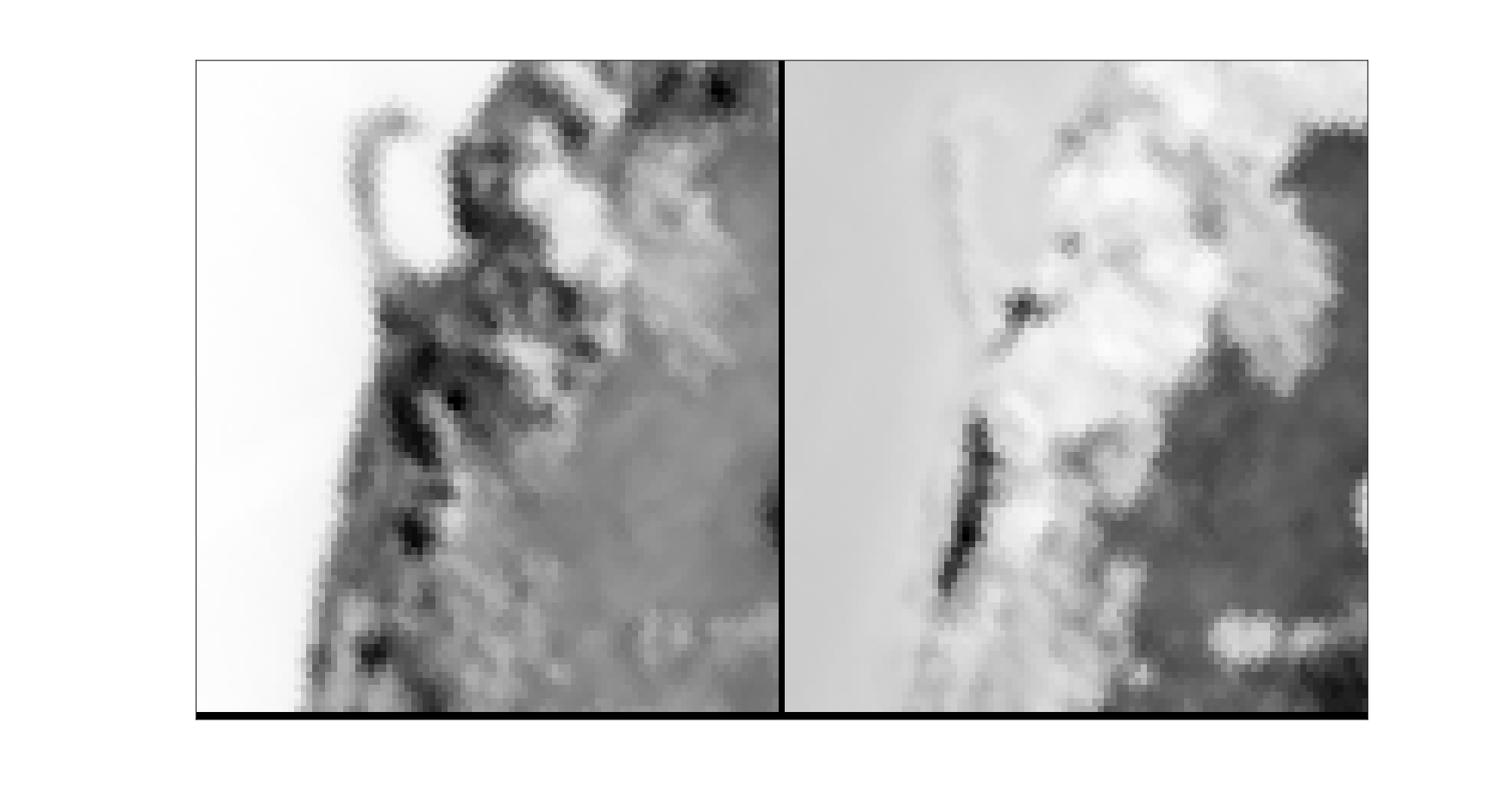} \\
(a) deep KL-NMF (Alg.~\ref{minvolADKL})
          & 
          (b) LC-DMF~\cite{de2023consistent} 
          & 
          (c) Multi-layer KL-NMF~\cite{cichocki2006multilayer, cichocki2007multilayer} 
         \end{tabular}
        \caption{Samson data set: From left to right abundance maps extracted from the second layer of min-vol deep KL-NMF (Alg.~\ref{minvolADKL}), LC-DMF~\cite{de2023consistent} and multi-layer KL-NMF~\cite{cichocki2006multilayer, cichocki2007multilayer}.}
        \label{fig:secondlayer_Samson}
\end{figure}

\newpage 

\section{Conclusion} 

In this paper, we have introduced deep NMF models based on $\beta$-divergences using the layer-centric loss function. 
We devised efficient multiplicative update  algorithms to estimate the parameters of these models. Our experimental results illustrated the practical efficacy of these approaches in diverse applications, namely facial feature extraction, topic modeling, and hyperspectral image unmixing. 

\revises{Future research directions include: 
\begin{itemize}
    \item exploring extrapolation strategies of BMMe introduced in~\cite{hien2024block}  to accelerate the convergence of the proposed algorithms (specifically, an extrapolation step such as $W^{ex}_\ell=W^{k-1}_\ell + \gamma_\ell^k [W^{k-1}_\ell - W^{k-2}_\ell]_+$, where $\gamma_\ell^k$ are extrapolation coefficients and $[\cdot]_+$ denotes the projection onto the nonnegative orthant, can be incorporated in the update of $W^{k-1}_\ell$, and similarly to $H_\ell^{k-1}$); 
    
    \item constructing a library of datasets and experiments  specifically designed for deep NMF models. This would facilitate more robust and
empirical evaluations, addressing this critical need in the field.

    \item applying deep NMF in other domains such as source separation and gene expression analysis.   
    
\end{itemize}  
}

  \section*{Acknowledgments}

\revises{The authors thank the reviewers for their feedback that helped improve the paper.}

 \appendix


\section{MU for deep $\beta$-NMF for 
$\beta = \{ 0, 1/2, 3/2 \}$} \label{sec:append} 

In this section, we derive MU  for $W$ in deep $\beta$-NMF (without regularization) for $\beta \in \{ 0, 1/2, 3/2\}$, as done for the KL divergence ($\beta = 1$) in Section~\ref{sec:algoNoregul}. 

\subsection{MU for $W$ when $\beta=3/2$}  

In order to derive MU to solving~\eqref{eq:probinhupper}, similarly to the case $\beta=1$, 
it is sufficient to solve~\eqref{eq:gradTocancel}. 
From Lemma~\ref{defG} and Table~\ref{table:conv_concav_decomp} for $\beta =3/2$, one can show that solving \eqref{eq:gradTocancel} reduces to finding  $w_{k_\ell}\geq 0$ of the following scalar equation:
\begin{equation} 
\label{beta3_2} a (w_{k_\ell})^{1/2} - b (w_{k_\ell})^{-1/2} -c=0,
\end{equation}
where $a=(\tilde w_{k_\ell})^{-1/2}\sum_{j_\ell} H_{k_{\ell}j}([\tilde{w}H]_{j_\ell})^{1/2}+2\lambda$, $b=(\tilde w_{k_\ell})^{1/2}\sum_{j_\ell} H_{k_{\ell}j_\ell} ([\tilde{w}H]_{j_\ell})^{-1/2} y_{j_\ell}$, $c=2\lambda (\bar w_{k_\ell})^{1/2}$ (note that we assume $\tilde w_{k_\ell}>0$ in Lemma~\ref{defG}). Equation \eqref{beta3_2} has the  positive solution 
$ 
w_{k_\ell} = \left(\frac{c+ \sqrt{c^2 + 4ab}}{2a}\right)^2.
$ 
In matrix form, we obtain the following multiplicative update 
$$W = 
1/4\left [ \frac{\Big[C+ \sqrt{C^{.2} + 4 A \odot B }\Big]}{[A]}\right]^{.2},
$$
where $A = \tilde W^{.(-1/2)}\odot  \left( \left[ \tilde{W}H \right]^{.1/2} H^\top \right) +2\lambda$, $B=\tilde W^{.(1/2)}\odot  \left( \frac{\left[ Y \right]}{\left[ \tilde{W}H \right]^{.1/2}}H^\top \right) $, and $C=2\lambda \Bar W^{.1/2}$. 
Similarly to the case $\beta=1$, the MU for $\beta=3/2$ would also encounter the zero locking phenomenon (as when $\tilde w_{k_\ell}=0$, we would have $b=0$).  This issue can be fixed by choosing an initial $W$ with strictly positive entries.  

\subsection{MU for $W$ when $\beta=0$}

From Lemma \ref{defG} and Table~\ref{table:conv_concav_decomp} for $\beta =0$, one can show that solving \eqref{eq:gradTocancel} reduces to finding  $w_{k_\ell}\geq 0$ of the following scalar equation:
\begin{equation} 
\label{beta0} a (w_{k_\ell})^{-2}+ \lambda (w_{k_\ell})^{-1}  -c=0,
\end{equation}
where $a=(\tilde w_{k_\ell})^2\sum_{j_\ell} H_{k_\ell j_\ell} \frac{y_{j_\ell}}{([\tilde w H]_{j_\ell})^2}$, $c=\sum_{j_\ell} \frac{H_{k_\ell j_\ell}}{[\tilde w H]_{j_\ell}}  + \frac{\lambda}{\bar w_{k_\ell}}$. Equation \eqref{beta0} has the following nonnegative solution: 
$
w_{k_\ell}=\frac{2a}{-\lambda+\sqrt{\lambda^2 + 4 ac}}
=\frac{\lambda+\sqrt{\lambda^2 + 4 ac}}{2c}.$
In matrix form, we obtain the following multiplicative update 
$$
W = 
1/2\frac{\Big[\lambda+\sqrt{\lambda^2 + 4 A\odot C} \Big]}{ [C]}, 
$$
where $A=\tilde W^{.2} \odot 
\left(\frac{Y}{[\tilde W H]^{.2}} H^\top\right)$ and $C=\frac{H^\top}{\tilde W H}+\frac{\lambda}{\Bar W} $.
We see that MU for the case $\beta=0$ also encounters the zero locking phenomenon (when $\tilde w_{k_\ell}=0$, we would have $a=0$), which can be fixed by choosing an initial $W$ with strictly positive entries.

\revises{\subsection{MU for $W$ when $\beta=\frac{1}{2}$}

From Lemma \ref{defG} and Table~\ref{table:conv_concav_decomp} for $\beta =\frac{1}{2} $, one can show that solving \eqref{eq:gradTocancel} reduces to finding  $w_{k_\ell}\geq 0$ of the following scalar equation:
\begin{equation} 
\label{beta1over1} -\Bar{c} (w_{k_\ell})^{\frac{3}{2}} + \Bar{b} (w_{k_\ell}) + \Bar{a} = 0 ,
\end{equation}
where $\Bar{a}=(\tilde w_{k_\ell})^{\frac{3}{2}}\sum_{j_\ell} H_{k_\ell j_\ell} \frac{y_{j_\ell}}{([\tilde w H]_{j_\ell})^{\frac{3}{2}}}$, $\Bar{c}=\sum_{j_\ell} \frac{H_{k_\ell j_\ell}}{([\tilde w H]_{j_\ell})^{\frac{1}{2}}}  + \frac{2\lambda}{(\bar w_{k_\ell})^{\frac{1}{2}}}$, and $\Bar{b}=2 \lambda$. Assuming $w_{k_\ell}$ nonnegative and posing $x=(w_{k_\ell})^{\frac{1}{2}}$,  Equation \eqref{beta1over1} can be equivalently rewritten as follows:
\begin{equation} 
\label{beta1over1_poly} x^3 + p x^2 + q x + r = 0 ,
\end{equation}
where $p=\frac{\Bar{b}}{-\Bar{c}}$, $q=0$ and $r=\frac{\Bar{a}}{-\Bar{c}}$ where $\Bar{c} > 0$ as we assume $\lambda > 0$ and $\bar w_{k_\ell} > 0$. 
Equation \eqref{beta1over1_poly} can be rewritten in the so-called normal form by posing $x = z - p/3$ as follows:
\begin{equation} 
\label{beta1over2_poly_2} z^3 + a z  + b = 0 ,
\end{equation}
where $a=\frac{1}{3}(3q - p^2)=\frac{-p^2}{3}$ and $b=\frac{1}{27}(2p^3 - 9pq + 27r)=\frac{1}{27}(2p^3 + 27r)$.
Under the condition $\frac{b^2}{4} + \frac{a^3}{27} > 0$, Equation \eqref{beta1over2_poly_2} has one real root and two conjugate imaginary roots. This condition boils down to $\frac{1}{4}(2p^3 + 27r)^2 - p^6 > 0$, which holds if $p \leq 0$ and $r<0$. By further assuming $\Bar{a} > 0$, these two conditions on $p$ and $r$ hold per the definitions of $p$ and $r$ given $\Bar{b}$ nonnegative and since $\Bar{c} >0$. Therefore, the positive real root has the following closed-form expression:
\begin{equation} 
\label{sol} z = \sqrt[3]{-\frac{b}{2} + \sqrt{\frac{b^2}{4} + \frac{a^3}{27}}} +  \sqrt[3]{-\frac{b}{2} - \sqrt{\frac{b^2}{4} + \frac{a^3}{27}}},
\end{equation}
and hence 
\begin{equation} 
\label{sol_2} w_{k_\ell} = \left(z + \frac{\Bar{b}}{3 \Bar{c}}\right)^2,
\end{equation} 
In matrix form, we obtain the following update 
$$
W = \left[ Z + \frac{[\Bar{B}]}{[3 \Bar{C} ]} \right]^{.2}, 
$$
where $Z=\Big[ \frac{-B}{2} + \Big[ \frac{[B]^{.2}}{4} + \frac{[A]^{.3}}{27} \Big]^{.(1/2)}\Big]^{.(1/3)} + \Big[ \frac{-B}{2} - \Big[ \frac{[B]^{.2}}{4} + \frac{[A]^{.3}}{27} \Big]^{.(1/2)}\Big]^{.(1/3)}$, $A=-\frac{1}{3} \Big[ \frac{[\Bar{B}]}{[-\Bar{C}]}\Big]^{.2}$, $B=\frac{1}{27} \Big( 2 \Big[  \frac{[\Bar{B}]}{[-\Bar{C}]}\Big]^{.3} + 27 \frac{[\Bar{A}]}{[-\Bar{C}]}\Big)$, $\Bar{B}= 2 \lambda$, $\Bar{C}=\frac{[E]}{[\Tilde{W} H]^{.(1/2)}} H^T + 2 \frac{\lambda}{[\Bar{W}]^{.(1/2)}}$, with $E$ denoting a matrix full of ones of appropriate size, and $\Bar{A}=[\Tilde{W}]^{.(3/2)} \odot \Big( \frac{[Y]}{[\Tilde{W}H]^{.(3/2)}} H^T \Big)$. 
}

\section{Facial feature extraction with deep KL-NMF} \label{sec:append_2}


We employ deep KL-NMF and compare its outcomes with multilayer KL-NMF. The testing methodology aligns with the approach outlined in Section \ref{subsec:facialExtraction}, with the only deviations being the selection of $\beta$, set to $\beta=1$, and the adjustment of the number of layers to three, with the ranks $r = [80, 40, 20]$.

\begin{figure}[ht!]
\begin{center}
\includegraphics[width=11cm]{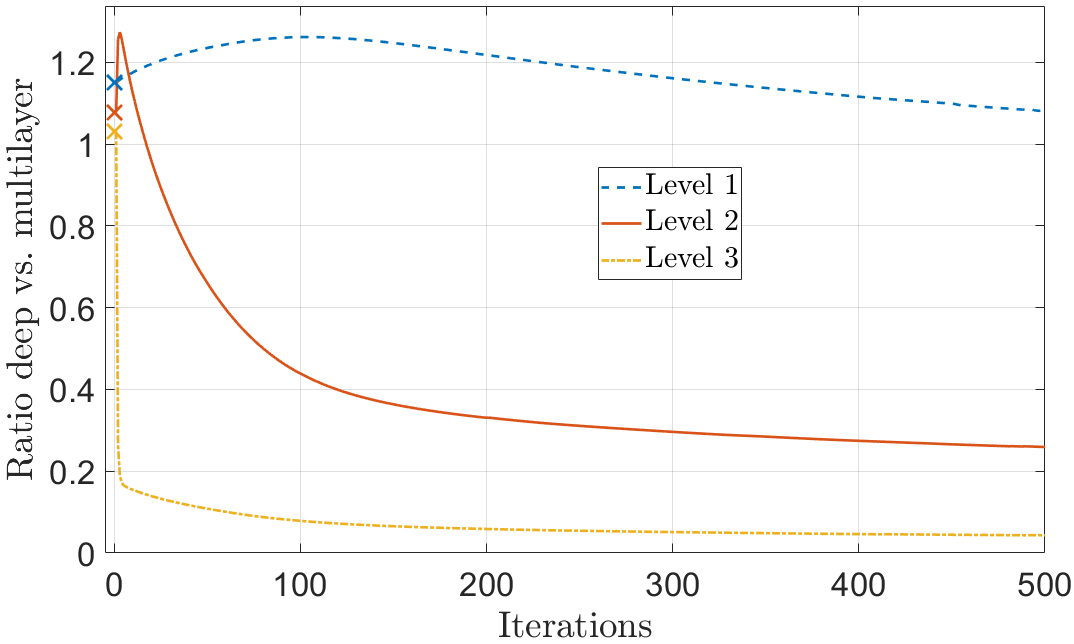} 
\caption{Evolution of the median errors at the different levels of deep KL-NMF (initialized with multilayer KL-NMF after 500 iterations) divided by the error of multilayer KL-NMF after 1000 iterations.} 
\label{fig:Msvsdeepnmfcbcl_KL}
\end{center}
\end{figure}




Figure~\ref{fig:Msvsdeepnmfcbcl_KL} depicts the median progression of the error across the three layers of deep KL-NMF (i.e., $D_{KL}(W_{\ell-1},W_\ell H_\ell)$), normalized by the final error achieved by multilayer KL-NMF after 1000 iterations. These ratios are initially larger than one since deep KL-NMF uses multilayer KL-NMF as initialization after only 500 iterations. 
As noted previously in the case of three layers with $\beta=\frac{3}{2}$, the error at the first layer remains larger than that of multilayer KL-NMF, as anticipated. However, the errors at the subsequent two layers rapidly become significantly smaller, consistent with observations in the first column of Table~\ref{tab:rescbcl_KL}.

\begin{center}
\begin{table}[ht!]
\begin{center} 
\begin{tabular}{|c|c|c|c|}
\hline 
   & error deep/ML (in \%) 
   & sparsity ML (in \%) 
   & sparsity deep (in \%)   \\  \hline
Layer 1 & 108.3 $\pm$   1.1  &   58.5  $\pm$  0.6 &  70.1  $\pm$  1.2 \\  
Layer 2 & 26.8  $\pm$   3.2  &    37.3  $\pm$  1.0            & 50.6  $\pm$  2.6    \\  
Layer 3 & 4.4   $\pm$   0.3  &         22.1  $\pm$  1.0          & 27.1 $\pm$   2.9 \\  
\hline
\end{tabular} 
\end{center}
\caption{Deep vs.\ Multilayer (ML) KL-NMF: average and standard deviation for the error of deep KL-NMF divided by that of ML KL-NMF (second column), and average and standard deviation for the sparsity of the facial features of ML  KL-NMF (second column) and  deep KL-NMF (third column). }
\label{tab:rescbcl_KL} 
\end{table}
\end{center}

\begin{figure}[ht!]
\begin{center}
\begin{tabular}{cc}
Multilayer - layer 1 & Deep - layer 1 \\
  \includegraphics[width=0.43\textwidth]{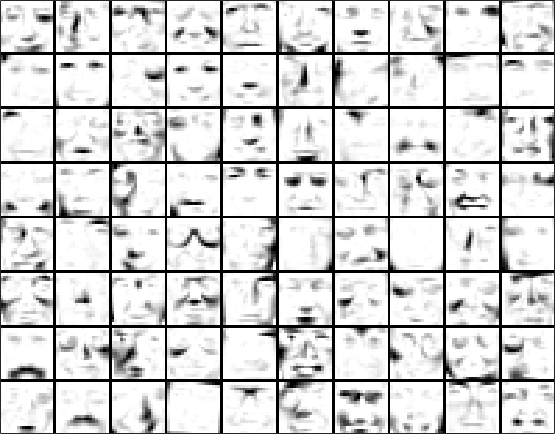} 
    &    
\includegraphics[width=0.43\textwidth]{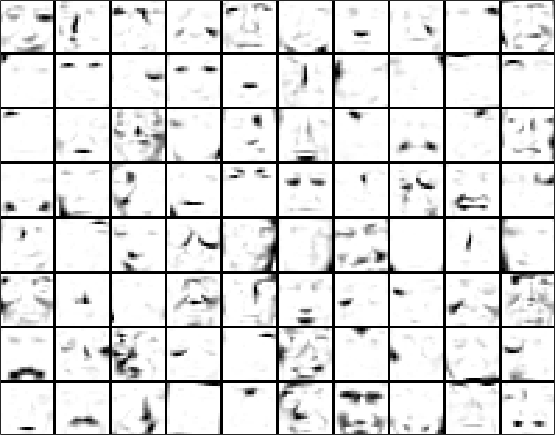} \\ 
Multilayer - layer 2 & Deep - layer 2 \\
  \includegraphics[width=0.43\textwidth]{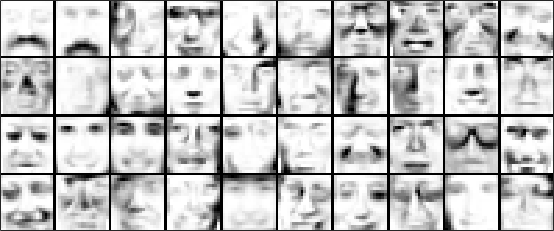} 
    &    
\includegraphics[width=0.43\textwidth]{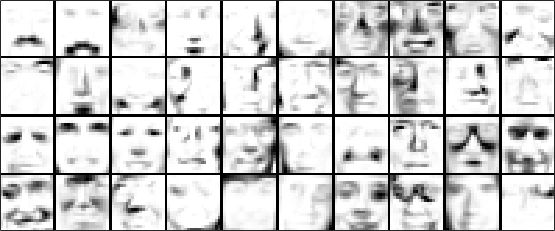} \\ 
Multilayer - layer 3 & Deep - layer 3 \\ 
  \includegraphics[width=0.43\textwidth]{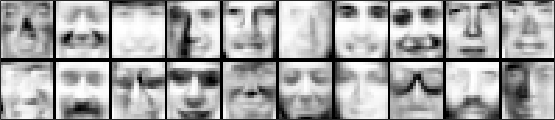} 
    &    
\includegraphics[width=0.43\textwidth]{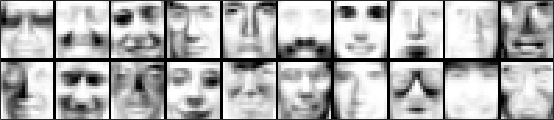} 
\end{tabular}
\caption{Example of facial features extracted by multilayer KL-NMF vs.\  deep KL-NMF. } 
\label{fig:Msvsdeepnmfcbclfacialfeatures_KL}
\end{center}
\end{figure}

Upon convergence, the error at the second (resp. third) layer is approximately four (resp. twenty) times smaller, on average, than that of multilayer KL-NMF. In both deep and multilayer KL-NMF, sparsity diminishes as the factorization progresses. Additionally, deep KL-NMF yields markedly sparser facial features, with an increase of 11.6\% at layer 1, 13.2\% at layer 2, and 5.1\% at layer 3. Facial features for multilayer and deep NMF at different layers (based on the last run of our experiment, as averaging facial features is nonsensical) are presented in Figure~\ref{fig:Msvsdeepnmfcbclfacialfeatures_KL}. Notably, most facial features in the first two layers of deep and multilayer KL-NMF exhibit similarity, differing primarily in sparsity. However, at the third layer, where the error of deep KL-NMF is significantly smaller, most facial features diverge substantially (e.g., the first one), illustrating that deep KL-NMF produces outcomes markedly distinct from multilayer KL-NMF.







\newpage 

\small 

\bibliographystyle{spmpsci}
\bibliography{Article2018}

\end{document}